\documentclass[accepted]{uai2021}
\usepackage{csquotes}
\usepackage[american]{babel}
\usepackage{url}
\usepackage{bm}
\usepackage[utf8]{inputenc}
\usepackage{amsmath}
\usepackage{graphicx}

\usepackage{enumitem}

\usepackage[nameinlink]{cleveref}

\usepackage{blindtext}
\usepackage{framed}
\usepackage{amssymb}
\usepackage{amsthm}
\usepackage{thmtools}
\usepackage{xcolor}

\usepackage{booktabs}
\usepackage{multirow}
\usepackage{longtable}
\usepackage{etoolbox,siunitx}
\robustify\bfseries
\usepackage{tikz}
\usetikzlibrary{bayesnet}
\usetikzlibrary{arrows}
\usetikzlibrary{shapes,decorations,arrows,calc,arrows.meta,fit,positioning}
\tikzset{
    -Latex,auto,node distance =1 cm and 1 cm,semithick,
    state/.style ={ellipse, draw, minimum width = 0.7 cm},
    point/.style = {circle, draw, inner sep=0.04cm,fill,node contents={}},
    bidirected/.style={Latex-Latex,dashed},
    el/.style = {inner sep=2pt, align=left, sloped}
}

\Crefformat{equation}{(#2#1#3)}
\Crefformat{figure}{Figure~#2#1#3}
\Crefname{example}{Example}{Examples}
\Crefname{lemma}{Lemma}{Lemmas}
\Crefname{cor}{Corollary}{Corollaries}
\Crefname{theorem}{Theorem}{Theorems}
\Crefname{assumption}{Assumption}{Assumptions}
\crefname{equation}{eq.}{eqs.}
\Crefname{equation}{Eq.}{Eqs.}
\Crefname{section}{Section}{Section}

\declaretheorem[style=plain,numberwithin=section,name=Theorem]{theorem}
\declaretheorem[style=plain,sibling=theorem,name=Lemma]{lemma}

\declaretheorem[style=definition,sibling=theorem,name=Definition]{definition}

\numberwithin{equation}{section}

\usepackage{spverbatim}

\usepackage{xr}
\makeatletter
\newcommand*{\addFileDependency}[1]{
  \typeout{(#1)}
  \@addtofilelist{#1}
  \IfFileExists{#1}{}{\typeout{No file #1.}}
}
\makeatother

\newcommand{\g}{\,\vert\,}	
\newcommand{\s}{\,;\,}

\newcommand{\E}[1]{\mathbb{E}\left[#1\right]}	
\newcommand{\EE}[2]{\mathbb{E}_{#1}\left[#2\right]}

\newcommand{\epehe}{\mathbf{\epsilon_{\text{PEHE}}}}

\DeclareMathOperator*{\argmin}{arg\,min}

\newcommand{\rmdo}{\mathrm{do}}	
\newcommand{\cdo}{\mathrm{do}}

\newcommand{\exclude}{\setminus}
\newcommand{\pa}{\mathrm{Pa}}
\newcommand{\qinv}{Q^{\mathrm{inv}}}
\newcommand{\qinvhat}{\hat{Q}^{\mathrm{inv}}}
\newcommand{\grad}{\nabla}

\newcommand{\given}{\mid}

\newcommand{\dist}{\ \sim\ }
\newcommand{\distiid}{\overset{\mathrm{iid}}{\dist}}

\usepackage[%
minnames=1,maxnames=99,maxcitenames=2,	
style=authoryear,
doi=false,	
url=false,	
giveninits=true,	
natbib,	
backend=bibtex,	
sorting=nyt	
]{biblatex}%

\renewbibmacro*{title}{%
  \ifthenelse{\iffieldundef{title}\AND\iffieldundef{subtitle}}	
    {}	
    {\ifthenelse{\ifentrytype{article}\OR\ifentrytype{inbook}%
      \OR\ifentrytype{incollection}\OR\ifentrytype{inproceedings}%
      \OR\ifentrytype{inreference}}	
      {\printtext[title]{%
        \printfield[sentencecase]{title}%
        \setunit{\subtitlepunct}%
        \printfield[sentencecase]{subtitle}}}%
      {\printtext[title]{%
        \printfield[titlecase]{title}%
        \setunit{\subtitlepunct}%
        \printfield[titlecase]{subtitle}}}%
     \newunit}%
  \printfield{titleaddon}}

\AtEveryBibitem{%
\ifentrytype{article}{	
    \clearfield{url}%
    \clearfield{urldate}%
    \clearfield{eprint}	
    \clearfield{eid}	
}{}	
\ifentrytype{book}{	
    \clearfield{url}%
    \clearfield{urldate}%
}{}	
\ifentrytype{collection}{	
    \clearfield{url}%
    \clearfield{urldate}%
}{}	
\ifentrytype{incollection}{	
    \clearfield{url}%
    \clearfield{urldate}%
}{}	
}	

\AtEveryBibitem{	
    \clearfield{pages}	
    \clearfield{review}%
    \clearfield{series}
    \clearfield{volume}	
    \clearfield{month}	
    \clearfield{eprint}	
    \clearfield{isbn}	
    \clearfield{issn}	
    \clearlist{location}	
    \clearfield{series}	
    \clearlist{publisher}	
    \clearname{editor}	
}{}

\UseRawInputEncoding
\title{Invariant Representation Learning for Treatment Effect Estimation}

 \author[1]{Claudia Shi}
 \author[2,3]{Victor Veitch}
  \author[1]{David M. Blei}
\affil[1]{Columbia University}
 \affil[2]{Google Research}
  \affil[3]{The University of Chicago}
 \date{}

\begin{document}
\maketitle
\begin{abstract}
The defining challenge for causal inference from observational data is the presence of `confounders', covariates that affect both treatment assignment and the outcome.
To address this challenge, practitioners collect and adjust for the covariates, hoping that they adequately correct for confounding. 
However, including every observed covariate in the adjustment runs the risk of including `bad controls', variables that \emph{induce} bias when they are conditioned on.
The problem is that we do not always know which variables in the covariate set are safe to adjust for and which are not.
To address this problem, we develop Nearly Invariant Causal Estimation (NICE).
NICE uses invariant risk minimization (IRM) \citep{arjovsky2019invariant} to learn a representation of the covariates that, under some assumptions, strips out bad controls but preserves sufficient information to adjust for confounding.
Adjusting for the learned representation, rather than the covariates themselves, avoids the induced bias and provides valid causal inferences.
We evaluate NICE on both synthetic and semi-synthetic data. When the covariates contain unknown collider variables and other bad controls, NICE performs better than adjusting for all the covariates.
Code is available at \href{https://github.com/claudiashi57/nice}{github.com/claudiashi57/nice}.
\end{abstract}

\section{Introduction}
Consider the following causal inference problem.

We want to estimate the effect of sleeping pills on lung disease using electronic health records, collected from multiple hospitals around the world. For each hospital $e$ and patient $i$, we observe whether the drug was administered $T^e_i$, the patient's outcome $Y^e_i$, and their covariates $X_i^e$, which includes comprehensive health and socioeconomic information.  The different hospitals serve different populations, so the distribution of the covariate $X^e$ is different across the datasets. But the causal mechanism between sleeping pills $T^e$ and lung disease $Y^e$ remains the same across hospitals.

The data in this example are observational. One challenge to causal inference from observational data is the presence of \textit{confounding variables} that influence both $T$ and $Y$ \citep{rosenbaum1983central, pearl2000causality}.
To account for confounding, we try to find them among the covariates $X$ and then adjust for them, e.g., using a method like G-computation \citep{robins1986new}, backdoor adjustment \citep{pearl2009causality}, or inverse propensity score weighting \citep{austin2011introduction}. The selected covariates are called the adjustment set.

To ensure that we have adjusted for all confounding variables, we might include every covariate in the adjustment set. However, naively adjusting for all covariates runs the risk of including ``bad controls'' \citep{ bhattacharya2007instrumental, pearl2009causality, cinelli2020making}, variables that \emph{induce} bias when they are adjusted for. In the example, a health condition caused by lung disease would be a bad control. It is causally affected by the outcome.

How can we exclude bad controls from the adjustment set?
One approach is to select confounders through a causal graph \citep{pearl2009causality}.
We ask a domain expert to construct a causal graph or a class of equivalent graphs. We then select the confounders for the causal adjustment. 
However, in practice, we may have thousands of covariates in the dataset. It may be too difficult to construct a graph with thousands of nodes. 

Another approach is to restrict the adjustment set to those that are known to be pre-treatment covariates \citep{rosenbaum2002overt, rubin2009should}. However, this approach can lead us to include covariates that are predictive of treatment assignment but not the outcome. If the record is sufficiently rich, this information can lead to near-perfect prediction of treatment, which is a problem for causal inference. Specifically, this creates an apparent violation of overlap, the requirement that each unit had a non-zero probability of receiving treatment \citep{d2020overlap}.
Practically, near-violations of overlap can lead to unstable or high-variance estimates of treatment effects \citep{ding2017instrumental}.

But these methods, and their challenges, suggest a new approach for causal estimation --- we want a representation of the covariates that contains sufficient information for causal adjustment, excludes bad controls, and helps provide low-variance causal estimates. This paper presents a method to find such a representation. 
\paragraph{Problem.}
We now state the problem plainly. 
We want to do causal inference with data collected from multiple environments, as in the hospitals' example above. The observed covariates are rich --- including all the causal parents of the outcome. There are no unobserved confounders, but identifiability \citep{pearl2000causality} or strong ignorability \citep{rosenbaum1983central} is \emph{not} guaranteed, due to the possible existence of bad controls. We do not know which covariates are safe to adjust for. The main question is: 
how can we use the multiple environments to find a representation of the covariates for valid causal estimation?

To address this question, we develop nearly invariant causal estimation (NICE), an estimation procedure for causal inference from observational data where the data comes from multiple datasets. The datasets are drawn from distinct environments, corresponding to distinct distributions of the covariates.

NICE applies Invariant Risk Minimization (IRM) \citep{arjovsky2019invariant} for causal adjustment. IRM is a framework for solving prediction problems. The goal is produce a predictor that is robust to changes in the deployment domain. The IRM procedure uses data from multiple environments to learn an \textit{invariant representation} $\Phi(T,X)$, a function such that the outcome $Y$ and the representation of the treatment and covariates $\Phi(T,X)$ have the same relationship in each environment. Predictors built on top of this representation will have the desired robustness. 

The main insight that enables NICE is that the IRM invariant representation also suffices for causal adjustment.
Informally, a representation is invariant if and only if it is informationally equivalent to the causal parents of the outcome $Y$ \citep{arjovsky2019invariant}.
For example, an invariant representation of the medical records will isolate the causal parents of lung disease. 
Assuming \emph{no mediators} --- variables on the causal path between the treatment and outcome --- in the covariate set, the causal parents of $Y$ constitute an adjustment set that suffices for causal adjustment, minimally impacts overlap, and that excludes all bad controls.
Hence, adjusting for an invariant representation is a safe way to estimate the causal effect.\footnote{To keep the exposition simple, we defer the discussion of mediators to the appendix.}

\paragraph{Contributions. } This paper 
develops NICE, an estimation procedure that leverages data from multiple environments to do causal inference.
It articulates the theoretical conditions under which NICE provides unbiased causal estimates and evaluates the method on synthetic and semi-synthetic causal estimation problems.

\section{Related Work}\label{sec:related-work}
Estimating the treatment effect from observational data consists of identification and estimation. 
The motivating problem is related to identification --- we do not know what covariates to adjust for.

In the introduction, we discussed two widely applied adjustment approaches: selecting covariates basing on a causal graph \citep{pearl2000causality} and restricting to covariates that are known to be pre-treatment \citep{rosenbaum2002overt, rubin2009should}. 
Another approach to select the adjustment set is through causal discovery.

Causal discovery methods aim to recover causal relationships or causal direction from data \citep{murphy1999modelling,spirtes2000constructing, shimizu2006linear, glymour2019review, shortreed2017outcome, peters2016causal,mooij2016distinguishing, heinze2018invariant}. 
In particular, NICE shares the same setup as invariance based causal discovery methods.
\citet{peters2016causal, heinze2018invariant, pfister2019invariant} leverage multiple environments to find the causal predictors of the target variable in the linear, non-linear, and time series settings.

Causal discovery assumes that the observed covariates correspond to well-defined variables in the causal graph (e.g., no measurement issues).
The representation learning approach of NICE does not require this assumption.
Further, even in the case where this assumption holds,
causal discovery methods are designed to conservatively select parents of $Y$. 
In practice, they often fail to select many actual parents.
In \Cref{exp:linear-synthetic}, we show that while the causal discovery method \citep{peters2016causal} is better at stripping out bad controls, it also discards confounders, which leads to poor estimation quality.

With identification, we can then estimate the treatment effect. There is extensive literature on different statistical estimators \citep{austin2011introduction,glynn2010introduction,vanderweele2011new,funk2011doubly}
and machine learning methods adapted for causal inference \citep{hill2011bayesian,athey2016recursive,beck2000improving, hartford2017deep, shalit2016estimating,louizos2017causal, yoon2018ganite, shi2019adapting}. 
All these estimators and methods assume identification and focus on improving the finite sample estimation quality. In contrast, NICE considers a setting where identification is not guaranteed. 

NICE uses the principle of invariance to solve a causal inference problem. A thread of related work uses the same principle to tackle different problems.

The principle of invariance is: if a relationship between $X$ and $Y$ is causal, then it is invariant to perturbations that changes the distributions of $X$. Conversely, if a relationship is invariant to many different perturbations, it's likely be causal \citep{haavelmo1943statistical, buhlmann2018invariance}.
This principle inspired a line of causality-based domain adaptation and robust prediction work.

\citet{rojas2018invariant} apply the idea for causal transfer learning, assuming the conditional distribution of the target variable given some subset of covariates is the same across domains. \citet{magliacane2018domain} relax that assumption. \citet{peters2016causal, heinze2018invariant} apply this principle for causal variable selection from multiple environments.
\citet{zhang2020domain} recast the problem of domain adaptation as a problem of Bayesian inference on the graphical models. \citet{arjovsky2019invariant} advocate a new generalizable statistical learning principle that is based on the invariant principle. \citet{rosenfeld2020risks} critically examined the generalizability of the proposed principle and its implementations.

These works focus of \emph{robust prediction}. NICE focuses \emph{causal estimation}. NICE is complementary as it studies the idea of applying domain adaptation methods for causal estimation. 
In particular, we focus on the application of IRM for treatment effect estimation.

\section{Nearly Invariant Causal Estimation}\label{sec:method}
\begin{figure}
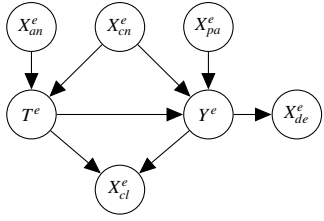

\centering
  \tikz{
 \node[latent,inner sep=.05cm,fill, minimum size=6.5mm] (x) {\scriptsize$X^e_{cn}$};%
 \node[latent,inner sep=.05cm,fill, minimum size=6.5mm, left=of x, xshift=0.5cm] (a) {\scriptsize$X^e_{an}$};%
 \node[latent,inner sep=.05cm,fill, minimum size=6.5mm, right=of x, xshift=-0.5cm] (b) {\scriptsize$X^e_{pa}$};%
 \node[latent,inner sep=.05cm,fill, minimum size=6.5mm,below=of a,yshift=0.5cm] (t) {\scriptsize$T^e$};
 \node[latent,inner sep=.05cm,fill, minimum size=6.5mm,below=of b ,yshift=0.5cm] (y) {\scriptsize$Y^e$}; 

 \node[latent,inner sep=.05cm,fill, minimum size=6.5mm, below=of x, yshift=-0.5cm] (c) {\scriptsize$X^e_{cl}$};%
 \node[latent,inner sep=.05cm,fill, minimum size=6.5mm,
 right=of y, xshift=-0.5cm] (d) {\scriptsize$X^e_{de}$};%
 \edge {x} {y,t};%
 \edge {t}{c, y};%
 \edge {a}{t};%
  \edge {b}{y};%
\edge {y}{d,c};%
 }
\caption{If the composition of $X^e$ is unknown, the treatment effect cannot be identified. (cn=confounder, cl=collider, pa=parents, an=ancestors, de=descendants)}\label{fig:complex_dag}
\end{figure}
We observe multiple datasets.  Each dataset is from an environment $e$, in which we observe a treatment $T^e$, an outcome $Y^e$, and other variables $X^e$, called covariates. Assume each environment involves the same causal mechanism between the causal parents of $Y^e$ and $Y^e$, but otherwise might be different from the others, e.g., in the distribution of $X^e$.  Assume we have enough information in $X^e$ to estimate the causal effect, i.e., it contains a set of variables sufficient for adjustment. But we do not know the status of each covariate in the causal graph.  A covariate might be an ancestor, confounder, collider, parent, or descendant. \Cref{fig:complex_dag} shows an example graph that defines these terms.

Each environment is a data generating process (DGP) with a causal graph and an associated probability distribution $P^e$.
The data from each environment is drawn i.i.d., $\{X^e_i, T^e_i, Y^e_i\} \distiid P^e$.
The causal mechanism relating $Y$ to $T$ and $X$ is assumed to be the same in each environment.
In the example from the introduction, different hospitals constitute different environments.
All the hospitals share the same causal mechanism for lung disease,
but they vary in the population distribution of who they serve, their propensity to prescribe sleeping pills,
and other aspects of the distribution.

The goal is to estimate the \emph{average treatment effect on
  the treated} (ATT)%
\footnote{For simple exposition, we focus on the ATT estimation. The method can also be applied to conditional average treatment effect or average treatment estimations.} in each environment,
\begin{align}
\begin{split}
  \label{eq:psi_do}
  \psi^e \triangleq \E{Y^e \g \rmdo(T^e=1), T^e=1} \\
  - \E{Y^e \g \rmdo(T^e=0), T^e=1}.
  \end{split}
\end{align}
The use of do notation \citep{pearl2000causality} indicates that the estimand is causal. 
The ATT is the difference between \emph{intervening} by assigning the treatment and intervening to prevent the treatment, averaged over the people who were actually assigned the treatment.
The causal effect for any given individual does not depend on the environment.
However, the ATT does depend on the environment because it averages over different populations of individuals.

\subsection{Causal estimation}
For the moment, consider one environment.  In theory, we can estimate the effect by adjusting for the confounding variables that influence both $T$ and $Y$ \citep{rosenbaum1983central}. Let $Z(X)$ be an \textit{admissible} subset of $X$---it contains no descendants of $Y$ and blocks all ``backdoor paths'' between $Y$ and $T$ \citep{pearl2014confounding}.
An admissible subset in \Cref{fig:complex_dag} is any that includes $X_{cn}$ but excludes $X_{cl}$ and $X_{de}$.  Using $Z(X)$, the causal effect can be expressed as a function of the observational distribution, 
\begin{align}
\begin{split}
  \label{eq:psi_adjusted}
  \psi & = \mathbb{E}_X[\EE{Y}{Y \g  T=1, Z(X)} \\
  & - \EE{Y}{Y \g T=0,  Z(X)}\g T=1].
\end{split}
\end{align}

We estimate $\psi$ in two stages. First, we fit a model $\hat{Q}$ for the conditional expectation $Q(T, Z(X)) = \EE{Y}{Y \g T, Z(X)}$.
Second, we use Monte Carlo to approximate the expectation over $X$,
\begin{align}
\label{eq:psi_hat}
   \hat{\psi} =   \frac{1}{\sum_i t_i} \sum_{i:t_i = 1}
    \left( \hat{Q}(1, Z(X_i)) - \hat{Q}(0, Z(X_i)) \right),  
\end{align}

The function $\hat{Q}$ can come from any model that predicts $Y$ from $\{T, Z(X)\}$.

If the causal graph is known then the admissible set $Z(X)$ can be easily selected and the estimation in \Cref{eq:psi_adjusted} is straightforward.  
But here we do not know the status of each covariate---if we inadvertently include bad controls in $Z(X)$ then we will bias the estimate.  
To solve this problem, we develop a method for learning an \textit{admissible representation} $\Phi(T,X)$, which is learned from datasets from multiple environments. 
An admissible representation is a function of the full set of covariates but one that captures the confounding factors and excludes the bad controls, i.e., the descendants of the outcome that can induce bias.\footnote{An admissible representation is analogous to an `admissible set' \citep{pearl2000causality}, which is a valid adjustment set.}  
Given the representation, we estimate the conditional expectations $\EE{Y}{Y \g \Phi(T,X)}$ and proceed to estimate the causal effect.

\subsection{Invariant Risk Minimization}
To learn an admissible representation, we use IRM. IRM is a framework for learning predictors that perform well across many environments.  We first review the main ideas of IRM and then adapt it to causal estimation.

Each environment is a causal structure and probability distribution.  Informally, for an environment to be valid, it must preserve the causal mechanism relating the outcome and the other variables.
\begin{definition}[Valid environment \citealp{arjovsky2019invariant}]\label{df:valid_intervention}
 Consider a causal graph $\mathcal{G}$ and a distribution $P(X, T, Y)$ respecting $\mathcal{G}$.
  Let $\mathcal{G}_e$ denote the graph under an intervention and $P^e = P(X^e, T^e, Y^e)$ be the distribution induced by the intervention. The intervention can be either atomic or stochastic. 
  An intervention is valid with respect to $(\mathcal{G}, P)$ if (i) $\EE{P^e}{Y^e| Pa(Y)}= \EE{P}{Y|Pa(Y)}$, and (ii) $V(Y^e |Pa(Y))$ is finite. An environment is valid with respect to $(\mathcal{G}, P)$ if it can be created by a valid intervention. 
\end{definition}

Given this definition, a natural notion of an invariant representation is one where the conditional expectation of the outcome is the same regardless of the environment.
\begin{definition}[Invariant representation]\label{df:invarianc-def}
  A representation $\Phi(T,X)$ is invariant with respect to environments $\mathcal{E}$ if and only if
  $\E{Y^{e_1}|\Phi(T^{e_1}, X^{e_1}) =\pi}= \E{Y^{e_2}|\Phi(T^{e_2}, X^{e_2})=\pi}$ for all $e_1, e_2 \in \mathcal{E}$.
\end{definition}

\Citet{arjovsky2019invariant} recast the problem of finding an invariant representation as one about prediction.  In this context, the goal of IRM is to learn a representation such that there is a single classifier $w$ that is optimal in all environments. Thus IRM seeks a composition $w \circ \Phi(T^e, X^e)$ that is a good estimate of $Y^e$ in the given set of environments. This estimate is composed of a representation $\Phi(T, X)$ and a classifier $w $ that estimates $Y$ from the representation.

\begin{definition}[Invariant representation via predictor \citealp{arjovsky2019invariant}]\label{df:irm-invariance-def}
  A data representation $\Phi: \mathcal{X} \rightarrow \mathcal{H}$ elicits an invariant predictor across environments $\mathcal{E}$ if there is a classifier $w : \mathcal{H} \rightarrow \mathcal{Y}$ that is simultaneously optimal for all environments. That is,
  \begin{align}
    \label{eq:irm-via-predict}
    w \in \argmin_{\bar{w}: \mathcal{H} \rightarrow \mathcal{Y}} R^{e}(\bar{w} \circ \Phi) \quad \textrm{for all \,} e \in \mathcal{E},
  \end{align} where $R^e$ is the the training objective's risk in environment $e$.
\end{definition}
The invariant representations in \Cref{df:invarianc-def,df:irm-invariance-def} align if we choose a loss function for which the minimizer of the associated risk in \Cref{eq:irm-via-predict} is a conditional expectation. (Examples include squared loss and cross entropy loss.)  
In this case, we can find an invariant predictor $ \qinv=  w  \circ \Phi(T^e, X^e)= \E{Y \g \Phi(T,X)}$ by solving \Cref{eq:irm-via-predict} for both $w$ and $\Phi$.

However, the general formulation of \Cref{eq:irm-via-predict} is computationally intractable, \citet{arjovsky2019invariant} introduce IRMv1 as a practical alternative.
\begin{definition}[IRMv1\citealp{arjovsky2019invariant} ]\label{df:irmv1}
IRMv1 is: 
\begin{equation}
  \hat{\Phi} = \argmin_{\Phi} \sum_{e \in \mathcal{E}} R^e(1.0 \cdot \Phi) +\lambda \parallel \grad_{w|w = 1.0} R^e(w \cdot \Phi) \parallel^2.
\end{equation}
\end{definition}
Notice here, IRMv1 fixes the classifier to the simplest possible choice: multiplication by the scalar constant $w=1.0$. 
The task is then to learn a representation $\Phi$ such that $w=1.0$ is the optimal classifier in all environments.
In effect, $\Phi$ becomes the invariant predictor, as $\qinv= 1.0 \cdot \Phi$.
The gradient norm penalizes model deviations from the optimal classifier in each environment $e$, enforcing the invariance.
The hyperparameter $\lambda$ controls the trade-off between invariance and predictive accuracy.\footnote{For details on IRMv1, see \citep[section 3.1]{arjovsky2019invariant}}

In practice, we parameterize $\Phi$ with a neural network that takes $\{t^e_i,x^e_i\}$ as input and outputs a real number. 
Let $\ell$ be a loss function, such as squared error or cross entropy, and $n_e$ be the number of units sampled in environment $e$.
Then, we learn $\hat{\Phi}$ by solving IRMv1 where each environment risk is replaced with the corresponding empirical risk:
\begin{equation}
\hat{R}^e(Q) = \frac{1}{n_e} \sum_{i} \ell(y^e_i, Q(t^e_i,x^e_i)).
\end{equation} $\qinvhat = 1.0 \cdot \hat{\Phi}$ is an empirical estimate of $\E{Y|\Phi(T, X)}$. 

\subsection{Nearly Invariant Causal Estimation}\label{sec:algorithm}
We now introduce nearly invariant causal estimation (NICE). NICE is a causal estimation procedure that uses data collected from multiple environments.
NICE exploits invariance across the environments to perform causal adjustment without
detailed knowledge of which covariates are bad controls.

Informally, the key connection between causality and invariance is
that if a representation is invariant across all valid environments then
the information in that representation is the information in the causal parents of $Y$.
Since the causal structure relevant to the outcome is invariant across environments, a representation
capturing only the causal parents will also be invariant. We can see that $\pa(Y)$ is the minimal
information required for invariance. A representation that is invariant over all valid environments will be minimal;
hence, an invariant representation must capture only the parents of $Y$.

NICE is based on two insights. First, as just explained, if $\Phi(T,X)$ is invariant over all valid environments, then $\E{Y | T, \pa(Y)\exclude \{T\}} = \E{Y | \Phi(T,X)}$.
Second, $\pa(Y) \exclude \{T\}$ suffices for causal adjustment. That is, $\pa(Y) \exclude \{T\}$ blocks any backdoor paths and does not include bad controls. Following \Cref{eq:psi_adjusted},

\begin{equation}
\begin{aligned}
  \psi &= \mathbb{E} [\E{Y \given T=1, \pa(Y)\exclude \{T\}}\\
  & - \E{Y \given T=0, \pa(Y)\exclude \{T\})} \g T=1]
  \end{aligned}
 \end{equation}
 Since $\E{Y | T, \pa(Y)\exclude \{T\}} = \E{Y | \Phi(T,X)}$,
 \begin{equation}
  \begin{aligned}
    \psi = \E{\E{Y \given \Phi(1,X)} - \E{Y \given \Phi(0,X)} \g T=1}.\label{eq:psi_phi}
  \end{aligned}     
 \end{equation}

Recall the invariant predictor $\qinv(T,X) = \E{Y \given \Phi(T,X)}$.
The NICE procedure is
\begin{enumerate}
\item Input: multiple datasets $\mathcal{D}_e := \{(X_i^e, Y_i^e, T_i^e)\}^{n_e}_{i=1}$.
\item Estimate the invariant predictor $\qinvhat = 1.0 \cdot \hat{\Phi}$ using an invariant objective, such as IRMv1. 
\item Compute $\hat{\psi}^e = \frac{1}{\sum_i t^e_i} \sum\limits_{i : t^e_i=1} \qinvhat(1,x^e_i)  - \qinvhat(0,x^e_i)$ for each environment $e$.
\end{enumerate}
Similar to the function $\hat{Q}$ in \Cref{eq:psi_adjusted}, $\qinvhat$ can come from any prediction model that uses an invariant objective. 
In \Cref{sec:experiment}, we use linear regression, TARNet \citep{shalit2016estimating}, and Dragonnet \citep{shi2019adapting}.

We call the procedure `nearly' invariant as we only ever have access to a limited number of environments, so we cannot be certain that we'll achieve invariance across all valid environments.

\section{Justification of NICE}\label{sec:theory}
We now establish the validity of NICE as a causal estimation procedure. All proofs are in the appendix.

First consider the case where we observe data from a sufficiently diverse set of environments that
the learned representation is invariant across all valid environments.
We prove that conditioning on a fully invariant representation is the same as conditioning on the
parents of $Y$.
\begin{lemma}\label{lem:generalizability}
Suppose that $\E{Y \g \pa(Y) = a} \neq \E{Y \g \pa(Y) = a'}$ whenever $a \neq a'$.Then a representation $\Phi$ is invariant across all valid environments if and only if $\E{Y^e |\Phi(T^e,X^e)} = \E{Y |\pa(Y)}$ for all valid environments.
\end{lemma}

\Cref{lem:generalizability} helps show that a representation that elicits an invariant predictor suffices for adjustment. 

\begin{theorem}\label{thm:identifiablity}
 Let $L$ be a loss function such that the minimizer of the associated risk is a conditional expectation, and let $\Phi$ be a representation that elicits a predictor $\qinv$ that is invariant for all valid environments. Assuming $X^e$ does not contain mediators between the treatment and the outcome, then
  $\psi^e = \E{\qinv(1,X^e) - \qinv(0,X^e)|T^e=1}$.
\end{theorem}

\Cref{thm:identifiablity} shows that the NICE estimand is equal to the ATT as long as the predictor $\qinv$ is invariant across all valid environments. 

In practice, if a predictor is invariant across a limited set of diverse environments, it may generalize to all valid environments.
Assuming a linear data generating process, \citet{arjovsky2019invariant} establish sufficient conditions on the number and diversity of the training environments such that the learned representation generalizes to all valid environments.
In the non-linear case, there are no known sufficiency results.
However, \citet{arjovsky2019invariant} give empirical evidence that access to even a few environments may suffice.\footnote{Establishing the sufficiency result of IRM is an open question. }

In addition to identifiability, non-parametric estimation of treatment effects with finite data, i.e., \Cref{eq:psi_hat}, requires `positivity' or `overlap' -- both treatment and non-treatment have a non-zero probability for all levels of the confounders \citep{rosenbaum1983central, imbens2004nonparametric}. Let $\Phi(X^e)$ be the covariate representation, i.e., $\Phi(X^e) = \{\Phi(T^e=1,X^e) , \Phi(T^e=0, X^e)\}$, in the following theorem, we establish that if the covariate set $X$ is sufficient for overlap, then $\Phi(X^e)$ is sufficient for overlap. 
\begin{theorem}\label{thm:overlap}
Suppose $\epsilon \leq P(T^e=1| X^e) \leq 1-\epsilon$ with probability 1, then $\epsilon \leq  P(T^e=1| \Phi(X^e)) \leq 1-\epsilon$ with probability 1.
\end{theorem}
\begin{figure}
\begin{minipage}{0.15\textwidth}
\centering
  \tikz{
 \node[latent,inner sep=.05cm, fill, minimum size=0.25pt] (x) {\scriptsize $X_1^e$};%
 \node[latent,inner sep=.05cm,below=of x,  fill, minimum size=0.25pt] (t) {\scriptsize $T^e$}; %
  \node[latent,inner sep=.05cm,right=of t, xshift=-0.5cm,fill, minimum size=0.25pt] (y) {\scriptsize $Y^e$}; %

\node[latent,inner sep=.05cm, right=of x, fill, xshift= -0.5cm, minimum size=0.25pt] (z) {\scriptsize $X_2^e$};%
 \edge {x} {y,t};%
 \edge {t} {y}
 }
 \caption*{(a) Noise}
\end{minipage}
\begin{minipage}{0.15\textwidth}
\centering
   \tikz{
 \node[latent,inner sep=.05cm, fill, minimum size=0.25pt] (x) {\scriptsize $X_1^e$};%
 \node[latent,inner sep=.05cm,below=of x,  fill, minimum size=0.25pt] (t) {\scriptsize $T^e$}; %
  \node[latent,inner sep=.05cm,right=of t, xshift=-0.5cm,fill, minimum size=0.25pt] (y) {\scriptsize $Y^e$}; %

\node[latent,inner sep=.05cm, right=of x, fill, xshift= -0.5cm, minimum size=0.25pt] (z) {\scriptsize $X_2^e$};%
 \edge {x} {y,t};%
 \edge {t} {y}
  \edge {y} {z}
 }
 \caption*{(b) Descendant}
\end{minipage}
\begin{minipage}{0.15\textwidth}
\centering
    \tikz{
 \node[latent,inner sep=.05cm, fill, minimum size=0.25pt] (x) {\scriptsize $X_1^e$};%
 \node[latent,inner sep=.05cm,below=of x,  fill, minimum size=0.25pt] (t) {\scriptsize $T^e$}; %
  \node[latent,inner sep=.05cm,right=of t, xshift=-0.5cm,fill, minimum size=0.25pt] (y) {\scriptsize $Y^e$}; %

\node[latent,inner sep=.05cm, right=of x, fill, xshift= -0.5cm, minimum size=0.25pt] (z) {\scriptsize $X_2^e$};%
 \edge {x} {y,t};%
 \edge {t} {y}
  \edge {t,y} {z}
 }
 \caption*{(c) Collider}
 \end{minipage}\hfill
\caption{We observe $\{X_1, X_2\}$, but do not know its composition. In (a) $\{X_1,X_2\}$ is a valid adjustment. In (b) and (c), $X_2$ is downstream of $Y$, so $\{X_1,X_2\}$ is not a valid adjustment.  }\label{fig:exp1}

\end{figure}
The intuition is that the richer the covariate set is, the more likely it is to  accurately predict the treatment assignment \citep{d2020overlap}. The covariate representation $\Phi(X^e)$, by definition, contains less information than $X^e$, therefore $\Phi(X^e)$ satisfies overlap if $X^e$ satisfies overlap.

Even when invariance across all valid environments is not guaranteed, NICE may still improve the estimation quality when there are possible colliders in the adjustment set. 
If the observed environments are induced by valid interventions, either atomic or stochastic, on the bad controls, an invariant representation over these environments can also exclude bad controls.
Even when the representation does not exclude the bad controls, invariance may remove at least some (if not all) collider dependence. Intuitively, conditioning on a subset of collider information should reduce bias in the resulting estimate. \Cref{thm:collider} in the appendix shows that this intuition holds for at least one illustrative causal structure. A fully general statement remains open.

\paragraph{The case of mediators.}
So far, we assumed the observed covariate set $X$ does not contain mediators between $T$ and $Y$. What happens to the interpretation of the learned parameter, $\hat{\psi}^e$, if the adjustment set contains mediators? 

Intuitively, NICE captures the information in the direct link between $T$ and $Y$. 
Concretely, if there are no mediators, the parameter reduces to ATT.
If there are mediators but no confounders, the parameter reduces to the Natural Direct Effect \citep{pearl2000causality}.
If there are mediators and confounders, we define the parameters as the natural direct effect on the treated (NDET).
The mathematics definitions are in the appendix. 
 
\section{Empirical Studies}\label{sec:experiment}
We study the performance of NICE with three experiments. We are interested in three empirical questions: 
(1) Does NICE strip out bad controls in practice?
(2) Is NICE "cost-less" when there are no bad controls?
(3) What is the effect of different amounts of environmental variation on NICE's performance.

We find that (1) when there are bad controls in the adjustment set, NICE can reduce bias induced by the bad controls. (2) When there are no bad controls in the adjustment set, NICE does not hurt the estimation quality. (3) Whether NICE can strip out bad controls depends on the diversity of the environments. The more diverse the environments, the more likely it is that NICE can strip out the bad controls.

\subsection{Experimental Setup}
We construct three experiments corresponding to different settings. 
We first consider the setting where NICE is theoretically guaranteed to strip out bad controls. 
In \Cref{exp:linear-synthetic}, the data are collected from diverse environments, and the DGPs are linear. 
In the non-linear setting, there are no known sufficiency results for the generalizability of IRM. Therefore, there is no theoretical guarantee that NICE can strip out bad controls. 

To study whether NICE can reduce bias from bad controls empirically, we validate NICE using non-linear semi-synthetic benchmark datasets in \Cref{exp:speed-dating}.
Furthermore, we study the effect of different amounts of environmental variation on NICE's performance in \Cref{exp:variation}.

\paragraph{Causal Estimands \&  Evaluation metrics.}
We consider two estimands: the sample average treatment effect on the treated (SATT), $\psi_s = \frac{1}{\sum_{i}t_i}\sum_{i:t_i=1} \left(Q(1,Z(x_i)) - Q( 0, Z(x_i))\right)$
and the conditional average treatment effect (CATE), $\tau(x_i) =Q(1,Z(x_i)) - Q(0, Z(x_i))$ \citep{imbens2004nonparametric}. 
For the SATT, the evaluation metric is the mean absolute error (MAE), $\epsilon_{att}= |\hat{\psi}_s - \psi_s|$. For the CATE, the metric is the Precision in Estimation of Heterogeneous Effect (PEHE) $\epehe = \frac{1}{n}\sum_0^n (\hat{\tau}(x_i) -\tau(x_i))^2$ \citep{hill2011bayesian}. PEHE reflects the ability to capture individual variation in treatment effects.
The main paper shows the MAE of the SATT averaged across environments. For the evaluation of CATE, see the appendix.
\begin{figure}
     \includegraphics[width=\linewidth]{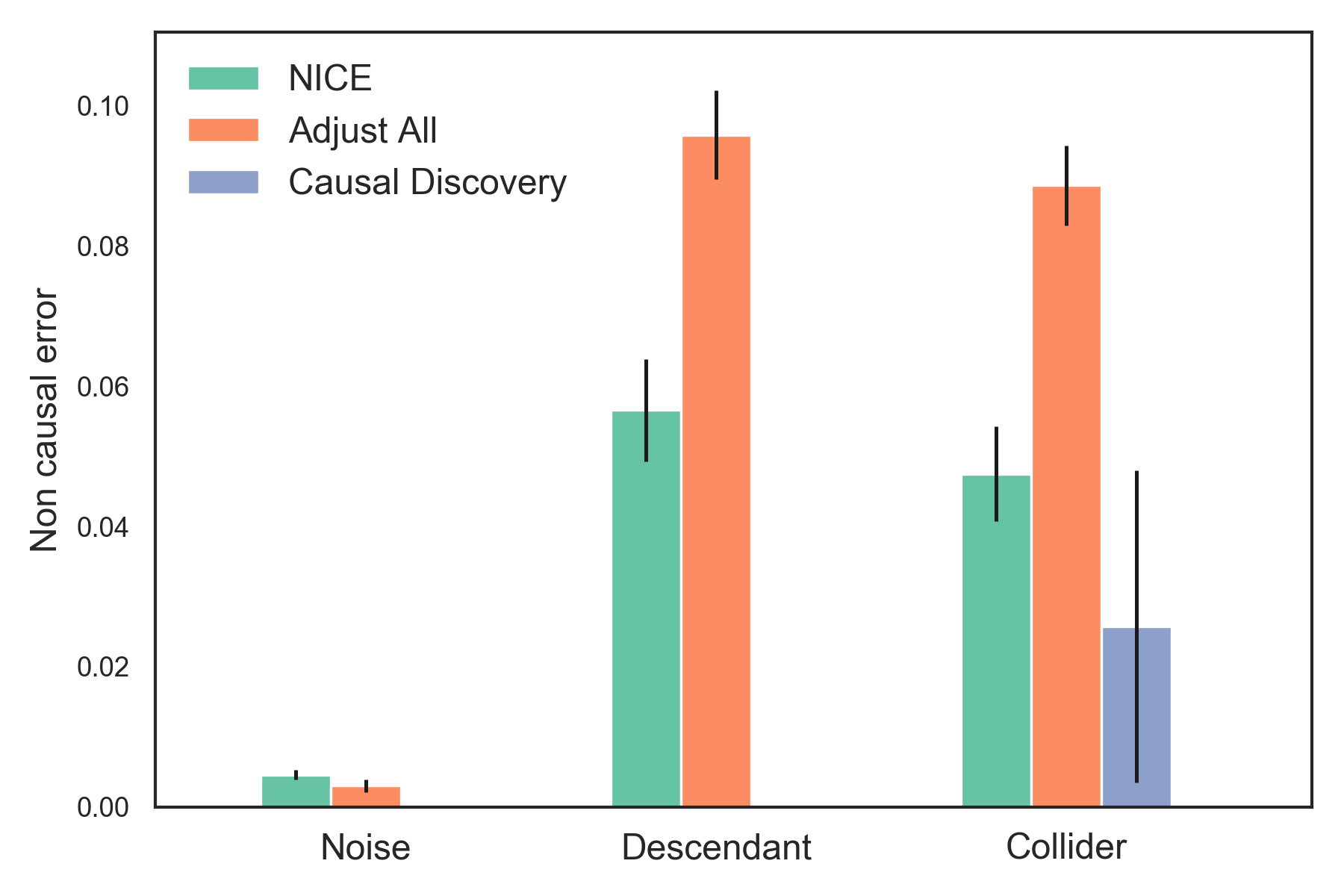}
    \caption{NICE strips out bad controls, which leads to better downstream treatment effect estimation. ICP (causal discovery) strips out bad controls, but also useful confounders (see \cref{fig:linear}). The non-causal error is measured by the mean square error of the weights on $X_2$. Lower is better. }
    \label{fig:linear-weight}
\end{figure}

\paragraph{Predictor Choices.}
Under the NICE procedure, the invariant predictor $\hat{Q}^{inv}$ can be any class of predictor trained with an IRMv1 objective. 
In the linear settings, we use OLS-2 as the predictor. 
OLS-2 is linear regression with two separate regressors for the treated and the control population. 

In the nonlinear settings, we consider two neural network models similar to the structure of TARNet \citep{shalit2016estimating} and Dragonnet \citep{shi2019adapting}. 
TARNet is a two-headed model with a shared representation $Z(X) \in R^p$, and two heads for the treated and control representation. 
The network has 4 layers for the shared representation and 3 layers for each expected outcome head. 
The hidden layer size is 250 for the shared representation layers and 100 for the expected outcome layers. 
We use Adam \citep{kingma2014adam} as the optimizer, set the learning rate as 0.001, and an l2 regularization rate of 0.0001. For Dragonnet, there's an additional treatment head, which makes treatment prediction from the shared representation.
For the hyper-parameter $\lambda$ used in IRMv1, we use $\lambda=10$ in the linear settings and $\lambda=100$ in the non-linear settings.

Since we use the same predictor across different DGPs, the hyper-parameters are chosen arbitrarily.
We use data from all environments to train and evaluate the predictor. 
The main paper presents results using TARnet. Results derived from Dragonnet are in the appendix. 

\begin{figure}
     \includegraphics[width=\linewidth]{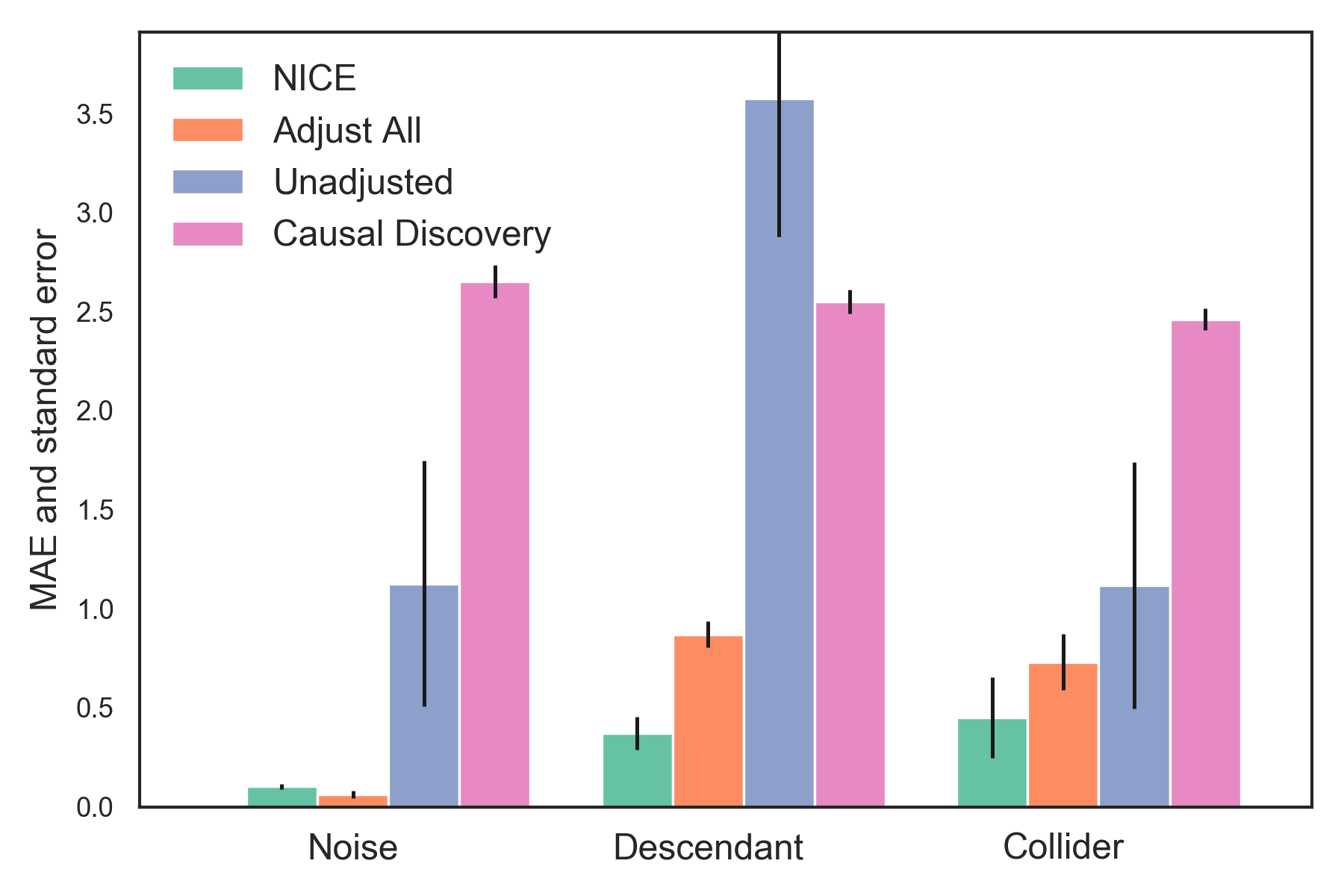}
    \caption{NICE reduces bias when the adjustment set contains bad controls and does not hurt if the adjustment set is valid. We use ICP for the causal discovery method, which is often too conservative. When ICP returns an empty set, estimated causal effect is zero. The figure reports average MAE and standard error of the SATT over 10 simulations.}
    \label{fig:linear}
\end{figure}

\paragraph{Adjustment Schemes.}
We compare estimation quality produced by the following adjustment schemes:
(1) adjusting for all covariates,
(2) NICE,
and (3) causal variable selection. 
Under (1), we pool the data across environments to fit a predictor $\hat{Q}$ and compute SATT using \Cref{eq:psi_adjusted}.
Under (3), we first use Invariant Causal Prediction (ICP) \citep{peters2016causal} to select an adjustment set. ICP is a variable selection method that identifies the target variable's causal parents by leveraging data from multiple environments.
We then pool data across environments, use the adjustment set to fit a predictor and compute SATT using \Cref{eq:psi_adjusted}.
The estimation procedure of NICE is described in \Cref{sec:algorithm}.

\subsection{NICE in linear settings}\label{exp:linear-synthetic}
We simulate data with the three causal graphs in \Cref{fig:exp1}. With a slight abuse of notation, each intervention $e$ generates a new environment $e$ with interventional distribution $P(X^e, T^e,  Y^e)$. $T^e$ is the binary treatment and $Y^e$ is the outcome. $X^e$ is a 10-dimensional covariate set that differs across DGPs. $X^e = (X^e_1, X^e_2)$, where $X^e_1$ is a five-dimensional confounder. $X^e_2$ is either noise, a descendant, or a collider in each DGP. The DGPs are:
\begin{align*}
X_1^e &\leftarrow \mathcal{N}(0, e^2)\\
T^e &\leftarrow Bern(\text{sigmoid} (X_1^e  \cdot w_{xt^e}+ \mathcal{N}(0, 1)) )\\
\tau &\leftarrow 5 + \mathcal{N}(0, \sigma^2) \\
Y^e &\leftarrow X_1^e \cdot w_{xy^e} +  T^e\cdot \tau + \mathcal{N}(0,e^2)
\end{align*}

In (a) $X_2^e \leftarrow \mathcal{N}(0, 1) $, in (b) $X_2^e \leftarrow e*Y^e + \mathcal{N}(0, 1)$, and in (c) $X_2^e \leftarrow e*Y^e + T^e + \mathcal{N}(0, 1)$.

For evaluation, following \citep{arjovsky2019invariant}, we create three environments $ \mathcal{E}=\{0.2, 2, 5\}$. We ran 10 simulations. In each simulation, we draw 1000 samples from each environment. We consider two types of variations: (1) whether the observed covariates $S(X)$ are scrambled versions of the true covariates $X$. If scrambled, $S$ is an orthogonal matrix. If not scrambled, $S$ is an identity matrix. 
(2) whether the treatment effects are heteroskedastic across environments. In the heteroskedastic setting $\tau \leftarrow 5 + \mathcal{N}(0, e^2)$. In the environment-level homoskedastic setting $\tau \leftarrow 5 + \mathcal{N}(0, 1)$.

We compare the estimation quality produced by four different adjustment approaches: 
(1) adjusting for all covariates,
(2) causal variable selection,
(3) NICE, and
(4) No adjustment.
The results in \Cref{fig:linear} and \Cref{fig:linear-weight} are under the unscrambled and heteroskedastic variant. The results of the other variants are in the appendix.

\paragraph{Analysis.} \Cref{fig:linear} reports the average of the MAE of SATT estimates over all three environments. 
We observe that when the covariate set does not include bad controls---simulation setting (a)--- NICE performs as well as adjusting for all covariates. 
When the covariate set includes bad controls that are closely related to the outcome, that is (b) and (c), 
NICE can help reduce the estimation bias. 

To understand why NICE reduces the estimation bias, we look at the weights of the control predictor. Ideally, the weights that correspond to the bad controls should be 0. As shown in \Cref{fig:linear-weight}, the predictor trained an IRMv1 objective places less weight on the bad controls than the predictor using empirical risk minimization. 
We observe ICP successfully strips out most of the bad controls. However, it produces worse causal estimates as it also strips out confounders. 
We believe that this is because (1) the amount of noise in the DGP is non-trivial, and (2) in some settings, the observed covariates are scrambled versions of the true covariates. The result suggests that while ICP is a robust causal discovery method, it should not be used for downstream estimation. 
A similar observation is made in \citet{zhaocomment}, where slight perturbations on ICP's assumptions might lead to poor performance.

\subsection{NICE in non-linear settings}\label{exp:speed-dating}
We validate NICE for the non-linear case on a benchmark dataset, SpeedDating. SpeedDating was collected to study the gender difference in mate selection \citep{fisman2006gender}. The study recruited university students to participate in speed dating, and collected objective and subjective information such as `undergraduate institution' and `perceived attractiveness'. It has 8378 entries and 185 covariates. ACIC 2019's simulation samples subsets of the covariates to simulate binary treatment $T$ and binary outcome $Y$. Specifically, it provides four modified DGPs: Mod1: parametric models; Mod2: complex models; Mod3: parametric models with poor overlap; Mod4: complex models with treatment heterogeneity. Each modification includes three versions: low, med, high, indicating an increasing number of covariates included in the models for $T$ and $Y$.

 \begin{table}[h]
\centering
\caption{
If the adjustment set is valid, NICE does not hurt the estimation performance. The table reports average MAE and bootstrap standard deviations of the SATT estimation.  }\label{tb:err_att}
\resizebox{\linewidth}{!}{
\begin{tabular}{ll|cccc}
\multicolumn{3}{l}{Valid Adjustment}  & \multicolumn{3}{c}{$\epsilon_{att}$} \\
\hline
& & {\textsc \textbf{Mod1}} &  {\textsc\textbf{Mod2} }     & {\textsc \textbf{Mod3}} & {\textsc \textbf{Mod4}}\\
\hline
\textbf{Low}  &Adjust All &$.04 \pm .08$  & $.05\pm .09$ & $.07 \pm .09$ & $.01 \pm .01$\\
& NICE &$.07 \pm .03$ & $.02 \pm .01$  & $.09\pm .03$ & $.04 \pm .02$ \\
\hline
\textbf{Med} & Adjust All &$.07 \pm .10$  & $.05\pm .05$ & $.04 \pm .04$ & $.07 \pm .08$\\
& NICE &$.05\pm .02$ & $.04 \pm .03$  & $.05 \pm .03$ & $.03 \pm .02$\\

\hline
\textbf{High} & Adjust All &$.07\pm .07$  & $.06\pm .05$ & $.06 \pm .07$ & $.04 \pm .04$\\
& NICE &$.02\pm .01$ & $.06 \pm .03$  & $.04\pm .02$ & $.07 \pm .04$\\
\hline
\end{tabular}}
\end{table}

\begin{table}[h]
\centering
\caption{NICE reduces estimation bias in the presence of bad controls. The table reports the average MAE and bootstrap standard deviation of SATT. }\label{tb:err_collider}
  \resizebox{\linewidth}{!}{
\begin{tabular}{ll|cccc}
\multicolumn{3}{l}{Bad Controls in Adjustment Set} & \multicolumn{3}{c}{$\epsilon_{att}$} \\
\hline
& & {\textsc \textbf{Mod1}} &  {\textsc\textbf{Mod2} }     & {\textsc \textbf{Mod3}} & {\textsc \textbf{Mod4}}\\
\hline
\textbf{low}  &Adjust All &$.26\pm .09$  & $.42\pm .03$ & $.34 \pm .08$ & $.46 \pm .09$\\
& NICE &$.09 \pm .07$ & $.03\pm .01$  & $.11 \pm .04$ & $.08 \pm .04$\\
\hline
\textbf{med}& Adjust All &$.38 \pm .10$  & $.35\pm .06$ & $.40 \pm .17$ & $.3\pm .09$\\
& NICE &$.06\pm .03$ & $.06 \pm .03$  & $.06 \pm .02$ & $.03 \pm .03$\\
\hline
\textbf{high} & Adjust All&$.32 \pm .14$  & $.38\pm .09$ & $.42 \pm .05$ & $.28\pm .05$\\
& NICE &$.05\pm .03$ & $.11 \pm .03$  & $.16\pm .05$ & $.11 \pm .05$\\
\hline
\end{tabular}}

\end{table}
The ACIC simulations are designed to assess the estimation quality of predictors and estimators. 
They do not come in multiple environments, nor do the covariates include bad controls.
To create multiple environments, we draw 6000 samples and select a covariate $x$ that's not the causal parent of $Y$. 
We sort the samples based on $x$ and divide them into three equal sized environments. 
For each DGP, we draw 10 bootstrap samples. 
To simulate bad controls, we included 20 copies of a collider in the adjustment set: $X^e_{co} = T^e + Y^e + \mathcal{N}(0, e^2)$, where $e \in \{0.01, 0.2, 1\}$.

\paragraph{Analysis. }
We compare two adjustment schemes: adjusting for all covariates and NICE. We first consider the setting where there are no bad controls. \Cref{tb:err_att} reports the average SATT MAE and standard deviations over 10 bootstraps under two adjustment schemes. 
We observe that NICE does not hurt the estimation quality in comparison to adjusting for all covariates. 

We also consider the setting where there is a strong collider. 
As shown in \Cref{tb:err_collider}, NICE reduces collider bias across simulation setups. However, we also observe that while it reduces the collider bias, it does not eliminate it completely. One potential reason is that the predictor is not optimal.

\subsection{The effect of environment variations on NICE's performance}\label{exp:variation}
\begin{figure}
\centering
\tikz{
 \node[latent, inner sep=.05cm,fill, minimum size=6.5mm] (at) {\scriptsize $A_t$};%

  \node[latent, inner sep=.05cm, right=of at,xshift=-0.5cm, fill, minimum size=6.5mm] (ay) {\scriptsize $A_y$};%
  \node[latent, inner sep=.05cm, right=of ay,xshift=-0.5cm,  fill, minimum size=6.5mm] (a1) {\scriptsize $A_1$};%
    
 \node[latent,inner sep=.05cm,below=of at, yshift= 0.5cm, fill, minimum size=6.5mm] (xt) {\scriptsize $X_t$}; %
  \node[latent,inner sep=.05cm,below=of ay, yshift= 0.5cm, fill, minimum size=6.5mm] (xy) {\scriptsize $X_y$}; %
   \node[latent,inner sep=.05cm,below=of xt, yshift= 0.5cm, fill, minimum size=6.5mm] (t) {\scriptsize $T$}; %
 \node[latent,inner sep=.05cm,below=of xy, yshift= 0.5cm, fill, minimum size=6.5mm] (y) {\scriptsize $Y$}; %
  \node[latent,inner sep=.05cm, right=of xy,  xshift=-0.5cm, fill, minimum size=6.5mm] (z) {\scriptsize $Z$};%
 \edge {at} {xt, z};%
  \edge {ay} {xy,z};%
   \edge {a1} {z};%

 \edge {xt} {y,t};%
 \edge {xy} {y};%
 \edge {t} {y};%
  \edge {y} {z};%
 }
 \caption{The DGP for \Cref{exp:variation}. The adjustment set \{X, A\} is valid. Adjustment set \{X, A, Z\} is not valid. }\label{fig:exp3-dgp}

\end{figure}
\begin{figure}
    \centering
    \includegraphics[width=\linewidth]{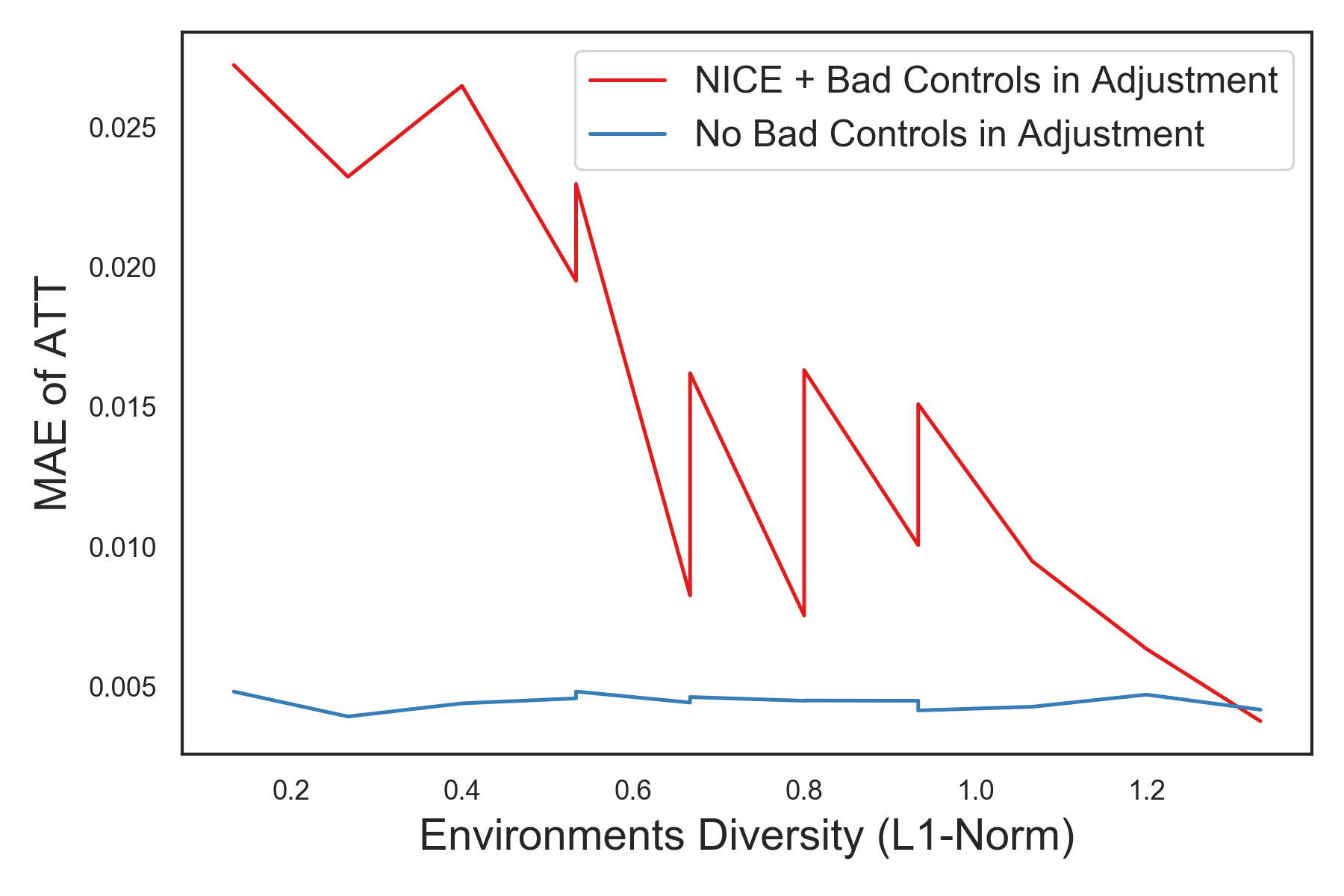}
    \caption{NICE mitigates bad controls more with access to more diverse environments.
    The x-axis is the environmental diversity. The y-axis is the average MAE of the SATT. }
    \label{fig:sample_porprotion}
\end{figure}

In this experiment, we examine the effect of environment variations on NICE's performance. 
We simulate non-linear data using the causal graph illustrated in \Cref{fig:exp3-dgp}.
The details of the data simulation are in the appendix.

We first draw three source environments $\{P^{e_1},P^{e_2}, P^{e_3}\}$ that are diverse.
To control the level of environment variation, we construct three new environments $\{P^{e'_1}, P^{e'_2}, P^{e'_3}\}$ that are mixtures of the three source environments.
Respectively, $P^{e'_1}, P^{e'_2}, P^{e'_3}$ draw ($p_1$, $p_2$, $p_3$) proportions from $P^{e_1}$, ($p_2$, $p_3$, $p_1$) proportions from $P^{e_2}$, and ($p_2$, $p_3$, $p_1$) proportions from $P^{e_3}$. The proportions $(p_1$, $p_2$, $p_3)$ sum to one. 

We approximate the diversity of the environments by the diversity of the proportions. The diversity measure is: $\frac{1}{{3}} \sum_{ij} |p_i- p_j|$. We consider 14 set of new environments, induced by different combination of the mixture probabilities.  We compare the estimation quality of NICE when given a covariate set that include bad controls $\{X, A, Z\}$ against adjusting for a valid covariate set $\{X, A\}$.

As shown in \Cref{fig:sample_porprotion}, the more diverse the environments, the more likely that NICE can strip out bad controls and reduce bias. When environments are sufficiently diverse, the learned representation is equivalent to a valid adjustment set.

\section{Discussion}
NICE lives at the intersection of representation learning and causal
inference, demonstrating how representation learning ideas can be
harnessed to improve causal estimation. Here we have examined the
causal setup where it's unknown which covariates are safe to adjust
for. One important direction for future work is to expand this setting to
one where we combine partial causal knowledge with representation
learning for estimating effects in more general scenarios.

\begin{acknowledgements}
This work is funded by ONR N00014-17-1-2131, 
ONR N00014-15-1-2209,
NIH 1U01MH115727-01,
NSF CCF-1740833,
DARPA SD2 FA8750-18-C-0130,
Amazon, and 
Simons Foundation. 
We thank Ryan Carey, Gemma Moran, Kevin Murphy, Dhanya Sridhar, and anonymous reviewers for helpful comments
and feedback. 

\end{acknowledgements}
\subsubsection*{References}
\printbibliography[heading=none]
\clearpage







\setcounter{section}{6}
\section{Appendix}
\makeatletter
\newtheorem{repeatthm@}{Theorem}
\newenvironment{repeatthm}[1]{%
    \def\therepeatthm@{\ref{#1}}
    \repeatthm@
}
{\endrepeatthm@}
\makeatother

\makeatletter
\newtheorem{repeatlem@}{Lemma}
\newenvironment{repeatlem}[1]{%
    \def\therepeatlem@{\ref{#1}}
    \repeatlem@
}
{\endrepeatlem@}
\makeatother

\makeatletter
\newtheorem{repeatprop@}{Proposition}
\newenvironment{repeatprop}[1]{%
    \def\therepeatprop@{\ref{#1}}
    \repeatprop@
}
{\endrepeatprop@}
\makeatother

\newtheorem{innercustomthm}{Corollary}
\newenvironment{customthm}[1]
  {\renewcommand\theinnercustomthm{#1}\innercustomthm}
  {\endinnercustomthm}

\subsection{Proofs of theorems}

\begin{repeatlem}{lem:generalizability}
  Suppose that $\E{Y \g \pa(Y) = a} \neq \E{Y \g \pa(Y) = a'}$ whenever $a \neq a'$.
  Then a representation $\Phi$ is invariant across all valid environments if and only if $\E{Y^e |\Phi(T^e,X^e)} = \E{Y |\pa(Y)}$ for all valid environments.
\end{repeatlem}
\begin{proof}
The if direction is immediate.

To establish the only if direction, we first show that $\Phi$ must contain at least $\pa(Y)$, in the sense $\E{Y \g \Phi(X)} = \E{Y \g \pa(Y)\cup Z}$ for some set $Z$.
We proceed with proof by contradiction.
Suppose that conditioning on $\Phi$ is equivalent to conditioning on only $\pa(Y) \exclude \{P\} \cup Z$, where $P$ is a parent of $Y$.
We now create two environments by setting $P=p$ and $P=p'$. Since $P$ is a parent of $Y$ this follows from the second rule of do calculus \citep{pearl2000causality}, 

\begin{align*}
  &\E{Y \g \pa(Y) \exclude \{P\} \cup Z; do(P=p)} \\
  &= \E{Y \g \pa(Y) \exclude \{P\} \cup Z, P=p}
\end{align*}

and
\begin{align*}
  &\E{Y \g \pa(Y) \exclude \{P\} \cup Z; do(P=p')} \\
  & = \E{Y \g \pa(Y) \exclude \{P\} \cup Z, P=p'}.
\end{align*}

The equality $\E{Y \g \pa(Y) \exclude \{P\}  \cup Z, P=p} = \E{Y \g \pa(Y) \exclude \{P\}  \cup Z, P=p'}$ holds only if $P$ is conditionally independent of $Y$ given $\pa(Y) \exclude \{P\} \cup Z$. Since $P$ is a parent of $Y$, by the first assumption of the lemma, the equality does not hold. It follows that  $\E{Y \g \pa(Y) \exclude \{P\} \cup Z \s do(P=p)} \neq \E{Y \g \pa(Y) \exclude \{P\} \cup Z \s do(P=p')}$. That is, if conditioning on $\Phi$ was equivalent to conditioning on less information than $\pa(Y) \cup Z$, then $\Phi$ would not be invariant across all valid environments.

It remains to show that $\Phi$ does not contain any more information than $\pa(Y)$. 

$\Phi$ cannot contain any descendants of the outcome. Suppose that $\Phi$ depends on some descendant $D$ of $Y$ in the sense that there is at least one environment and $d \neq d'$ where $\E{Y \g \Phi(X \exclude D, D=d)} \neq \E{Y \g \Phi(X \exclude D, D=d')}$. Then, construct a new environment $e$ by randomly intervening and setting $\cdo(D=d)$ or $\cdo(D=d')$, each with probability $0.5$. In this new environment, there is no relationship between $Y$ and $D$. Accordingly, $\E{Y^e \g \Phi(X^e \exclude D^e, D^e=d)} = \E{Y^e \g \Phi(X^e \exclude D^e, D^e=d')}$. 
Thus, the conditional expectations are not equal (as functions of $d$) in the two environments---a contradiction. 

Next, we show that, $\Phi$ needs not contain the non-parent ancestors $A$ of the outcome, because $\E{Y \g \{A\} \cup \pa(Y)} = \E{Y \g \pa(Y)}$ by
the Markov property of the causal graph, where $A$ is any non-ancestor variables.
Since $\Phi$ contains $\pa(Y)$, it follows that $\Phi$ does not depend on any non-parent ancestor $A$.

For expository purposes, the proof is done with $\cdo$ calculus \citep{pearl2000causality} for atomic interventions. If the environments are generated with stochastic interventions, we can use the same proof strategy with $\sigma$ calculus \citep{correa2020calculus}.

\end{proof}
\begin{repeatthm}{thm:identifiablity}
 Let $L$ be a loss function such that the minimizer of the associated risk is a conditional expectation, and let $\Phi$ be a representation that elicits a predictor $\qinv$ that is invariant for all valid distributions. Assuming there is no mediators between the treatment and the outcome, then
  $\psi^e = \E{\qinv(1,X^e) - \qinv(0,X^e)|T^e=1}$.
\end{repeatthm}

\begin{proof}
 We assume the technical condition of \Cref{lem:generalizability}, that $\E{Y^e \g \pa(Y^e) = a} \neq \E{Y^e \g \pa(Y^e) = a'}$ whenever $a \neq a'$. This is without loss of generality because violations of this condition will not lead to different causal effects.

  By the assumption on the loss function, the elicited invariant predictor is $\E{Y^e \given \Phi(T^e,X^e)}$.
  \Cref{lem:generalizability} shows that $\E{Y^e \given \Phi(T^e,X^e)} = \E{Y^e \given \pa(Y^e)} $. We further observe that the non-treatment parents of $Y^e$ are sufficient to block backdoor paths. It follows the ATT can be expressed as the following. 
  \begin{align*}
  \psi^e & =  \mathbb{E}[\E{Y^e \given T^e=1, \pa(Y^e)\exclude \{T^e\}} \\
  & - \E{Y^e \given T^e=0, \pa(Y^e)\exclude \{T^e\})} \g T^e=1]\\
  & = \E{\E{Y^e \given \Phi(1,X^e)} - \E{Y^e \given \Phi(0,X^e)} \g T^e=1} 
\end{align*}
\end{proof}

\begin{repeatthm}{thm:overlap}
Suppose $\epsilon \leq P(T^e=1| X^e) \leq 1- \epsilon$ with probability 1, then $\epsilon \leq P(T^e=1| \Phi(X^e)) \leq 1- \epsilon$ with probability 1.
\end{repeatthm}

\textit{Proof. } The proof follows directly from Theorem 1 in \cite{d2020overlap}. The intuition is that the richer the covariate set is, the more likely it is to predict the treatment assignment accurately \citep{d2020overlap}. The covariate representation $\Phi(X^e)$ by definition contains less information than $X^e$, therefore $\Phi(X^e)$ satisfies overlap if $X^e$ satisfies overlap.
\subsection{The case of Colliders}

\begin{figure}
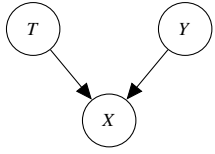

\centering
\tikz{
 \node[latent, fill] (x) {\scriptsize $X$};%
 \node[latent,above=of x,xshift=-1cm, yshift=-0.5cm , fill] (t) {\scriptsize $T$}; %
 \node[latent,above=of x,xshift=1cm , yshift=-0.5cm , fill] (y) {\scriptsize $Y$}; %
 \edge {y, t} {x};%
 }
        
  \caption{V-structure graph. We denote the bias induced by conditioning on $X$ as V-bias.  }\label{fig:thm-collider}

\end{figure}

Consider the DGP with binary variables $\{X, Y, T\}$ illustrated in \Cref{fig:thm-collider}, where $X$ is causally influence by $Y$ and $T$. 
\begin{theorem}\label{thm:collider}

  Let $cov$ denote the covariance between two variables, we define collider bias at $X = c$ as $\Delta(X =c)= cov(T,Y|X = c) - cov(T,Y)$, and collider bias of $X$
  as $\Delta(X) = |P(X=1) \Delta(X =1) +  P(X=0) \Delta(X =0)|$.
  Let $\Phi(T,X)$ be a random variable, where $P(\Phi(T,X) = X) \geq 0.5$.
  Suppose $P(X=1) =0.5$, and $\Delta(X=1)$ has the same sign as $\Delta(X=0)$, conditioning on $X$ induce more collider bias than conditioning its coarsening $\Phi(T,X)$:
\begin{center}
$\Delta(\Phi(T,X)) \leq \Delta(X)$
\end{center}
\end{theorem}

\begin{figure}
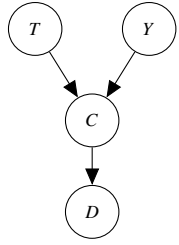

\centering
\tikz{
 \node[latent, fill] (x) {\scriptsize $C$};%
  \node[latent, fill, below=of x, yshift=0.5cm] (d) {\scriptsize $D$};%
 \node[latent,above=of x,xshift=-0.75cm, yshift= -0.5cm,fill] (t) {\scriptsize $T$}; %
 \node[latent,above=of x,xshift=0.75cm , yshift= -0.5cm,fill] (y) {\scriptsize $Y$}; %
 \edge {y, t} {x};%
  \edge {x} {d};%
 }
        
  \caption{Y-structure graph. We denote the bias induced by conditioning on $D$ as Y-bias. }\label{fig:thm-collider-2}
\end{figure}

\textit{Proof. } The proof follows corollary 2.1 in \cite{nguyen2019magnitude}. 
\begin{customthm}{2.1}
We refer to collider bias in the $V$ substructure embedded in the Y structure as `embedded V-bias' and denote it as $\Delta(C=c)$. For the covariance effect scale, Y-bias $\Delta(D=d)$ relates to embedded V-bias through the following formula:
\begin{align*}
    &\Delta(D=d)\\
    & = \frac{ p(D=d\g C=1)- p(D=d\g C=0)}{\{\mathrm{P}(D=d)\}^2}\\
    &\cdot \begin{bmatrix}
    p(D=d\g C=1)\{\mathrm{P}(C=1)\}^2\cdot \Delta(C=1)-\\
    p(D=d\g C=0)\{\mathrm{P}(C=0)\}^2\cdot \Delta(C=0)
    \end{bmatrix}.
\end{align*}
\end{customthm}
With the corollary above, let $D$ denote $\Phi(T,X)$, let $C$ denote the collider $X$ in \Cref{fig:thm-collider}. The bias induced by conditioning on $D$ is less than the bias induced by conditioning on $C$.
\begin{align*}
\Delta(D=1)
& = \frac{2\alpha - 1}{0.25}(0.25\alpha \cdot \Delta(C=1) \\
& -0.25(1-\alpha)  \cdot \Delta(C=0))\\
& = (2\alpha - 1)(\alpha \cdot \Delta(C=1) \\
& -(1-\alpha) \cdot  \Delta(C=0))\\
\Delta(D=0) 
& = \frac{1- 2\alpha}{0.25}(0.25(1-\alpha)\cdot \Delta(C=1) \\
& -0.25\alpha  \cdot \Delta(C=0))\\
& = ( 1- 2\alpha )((1-\alpha)\cdot \Delta(C=1) \\
& -\alpha \cdot  \Delta(C=0))\\
\Delta(C) &= |0.5\cdot  \Delta(C=0) +  0.5\cdot \Delta(C=1)|\\
\Delta(D) &= |0.5 \cdot \Delta(D=0) +  0.5\cdot \Delta(D=1)|\\
\Delta(D) &= |0.5(2 \alpha -1)^2 \cdot \Delta(C=1) \\
& + 0.5(2 \alpha -1)^2 \cdot \Delta(C=0)|\\
&\leq \Delta(C)
\end{align*}

\clearpage
\subsection{The Case of Mediators}
In the main paper, we assumed the covariate set $X$ contains no mediators between treatment and outcome. What happens to the interpretation of the learned parameter if the adjustment set contains mediators? Intuitively, NICE retains the direct link between the treatment and the outcome. Specifically, if there are no mediators, the parameter reduces to ATT. If there are mediators but no confounders, the parameter reduces to the Natural Direct Effect \citep{pearl2000causality}. If there are mediators and confounders, the NICE estimand is a non-standard causal target that we call the natural direct effect on the treated (NDET).

Conceptually, NDET describes the expected change in outcome $Y$ for the treated population, induced by changing the value of $T$, while keeping all mediating factors $M$, constant at whatever value they would have obtained under $\cdo(t)$.
The main point is that NDET provides answers to questions such as, ``does this treatment have a substantial direct effect on this outcome?''. Substantively, NDET is the natural direct effect, adjusted for confounders.

Formally, NDET for environment $e$ is 
\begin{equation}\label{eq:ndet}
    \begin{aligned}
    \psi^e &= \mathbb{E}_{M^e \g T^e=1}[\E{Y^e \g M^e \s \cdo(T^e=1)} \\
    &- \E{Y^e \g M^e \s \cdo(T^e=0)} |T^e=1].
    \end{aligned}
\end{equation}
With adjustment set $W^e$, the causal effect can be expressed through a parameter of the observational distribution:
\begin{equation}\label{eq:ndet_stats}
\begin{aligned}
\psi^e &= \mathbb{E}_{M^e, W^e}[\E{Y^e \g T^e=1,M^e, W^e} \\
& - \E{Y^e \g T^e=0,M^e, W^e}\g T^e=1].
\end{aligned}
\end{equation}

Importantly, the mediators $M^e$ and the confounders $W^e$ show up in the same way in \Cref{eq:ndet_stats}. Accordingly, we don't need to know which observed variables are mediators and which are confounders to compute the parameter.
Under the NICE procedure, we condition on all parents of $Y^e$, including possible mediators. Thus, the NICE estimand is the NDET in each environment.
\subsection{Details of the experiments}
\subsubsection{Experiment 1}
\begin{figure}
     \centering
     \includegraphics[width=\linewidth]{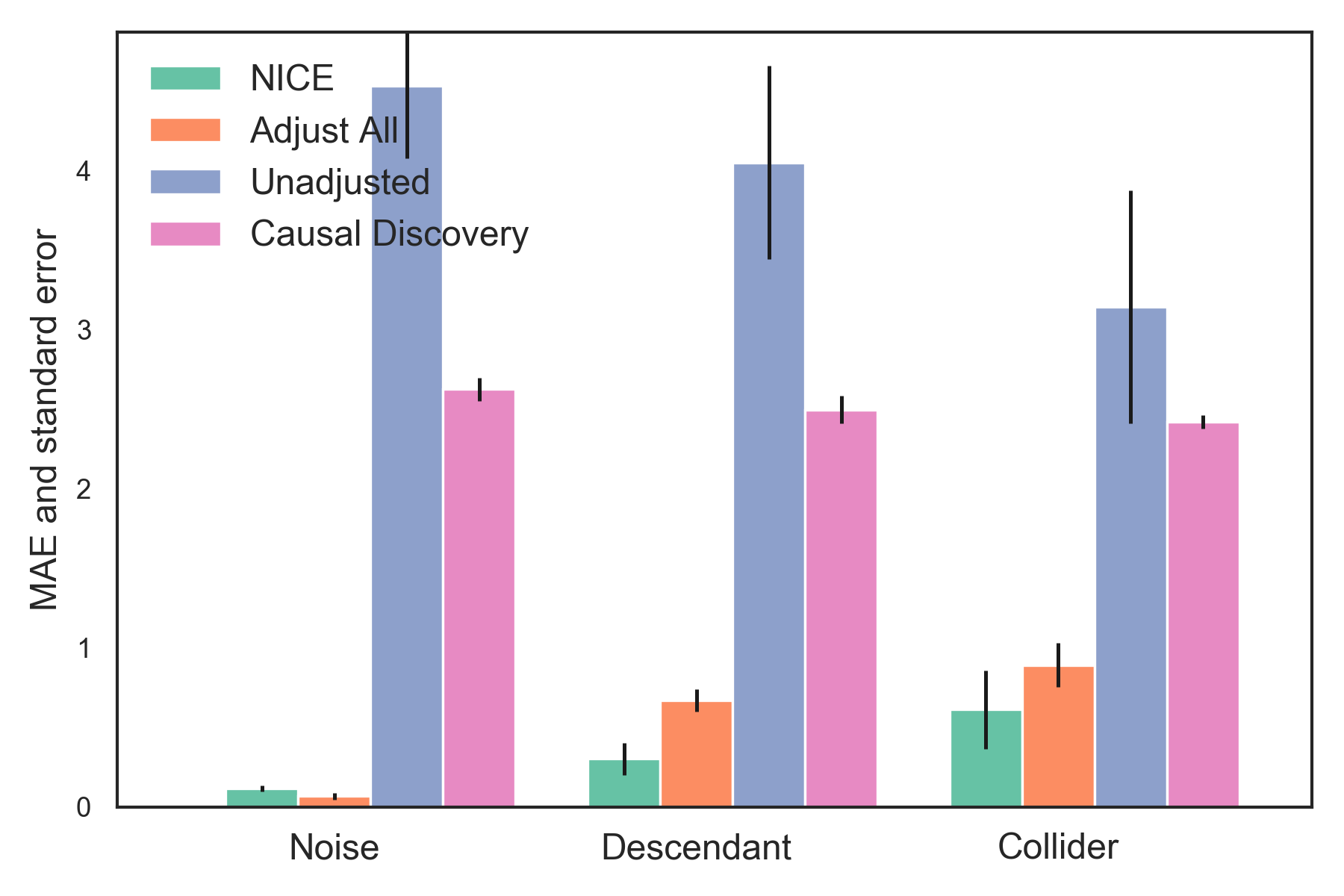}
    \caption{models performance under the scrambled and heteroskedastic setting }\label{fig:exp-res-1}
\end{figure}

\begin{figure}
      \centering
    \includegraphics[width=\linewidth]{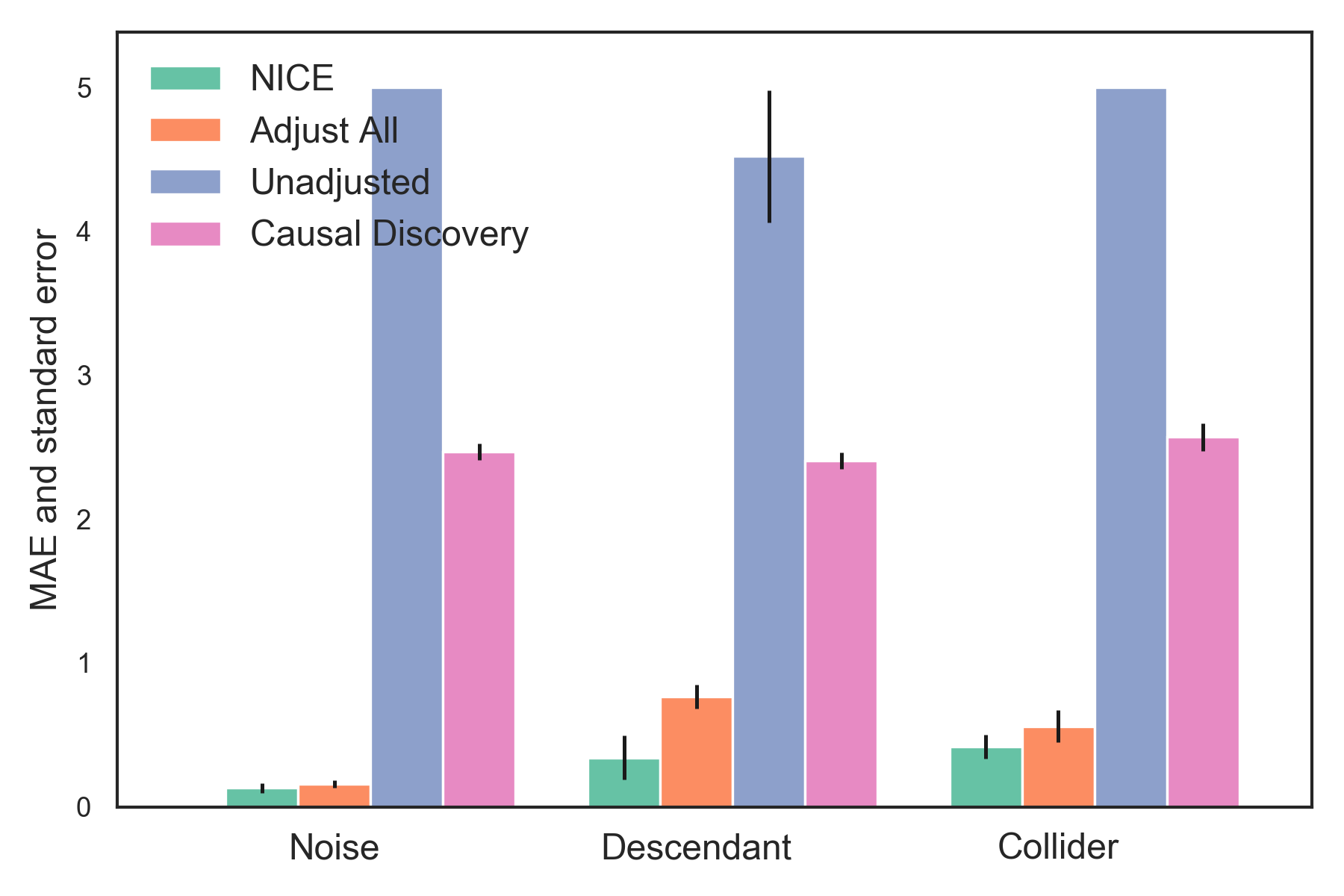}
    \caption{models performance under the scrambled and homoscedastic setting }\label{fig:exp-res-2}
\end{figure}

\begin{figure}
     \centering
     \includegraphics[width=\linewidth]{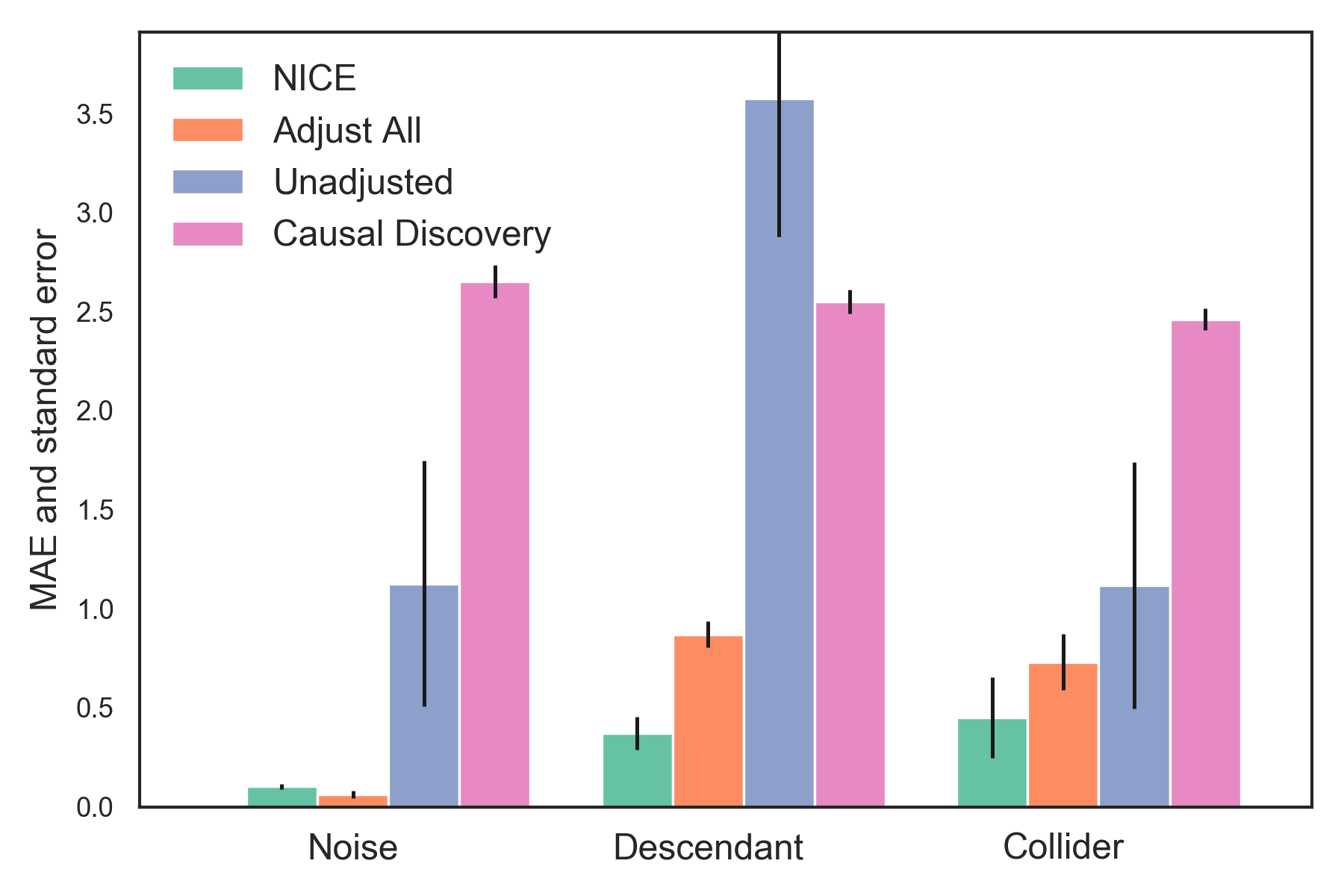}
    \caption{models performance under the unscrambled and heteroskedastic setting }\label{fig:exp-res-3}
\end{figure}

\begin{figure}
\centering
    \includegraphics[width=\linewidth]{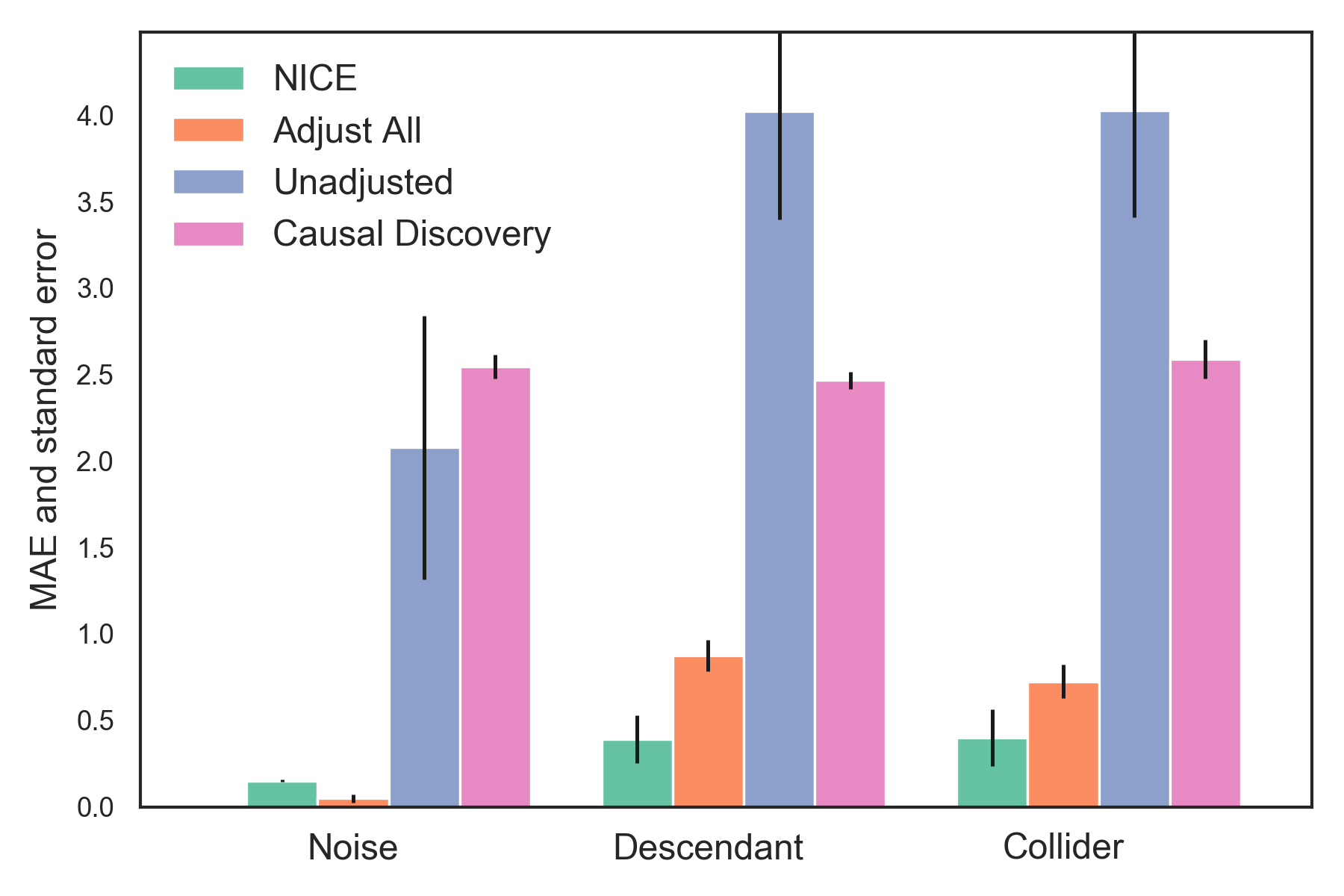}
\caption{models performance under the unscrambled and homoscedastic setting }\label{fig:exp-res-4}

\end{figure}

We evaluate the treatment effect estimation of various adjustment schemes using four variants of the data simulations. The variants are generated according to the following: 1) The observed covariates $S(X)$ are scrambled versions of the true covariates $X$. If scrambled, $S$ is an orthogonal matrix. If not scrambled, $S$ is an identity matrix. 
2) In the heteroskedastic setting $\tau \leftarrow 5 + \mathcal{N}(0, e^2)$. In the environment-level homoskedastic setting $\tau \leftarrow 5 + \mathcal{N}(0, 1)$.
The results are illustrated in \Cref{fig:exp-res-1}, \Cref{fig:exp-res-2}, \Cref{fig:exp-res-3}, and \Cref{fig:exp-res-4}.


\begin{figure}
     \centering
     \includegraphics[width=\linewidth]{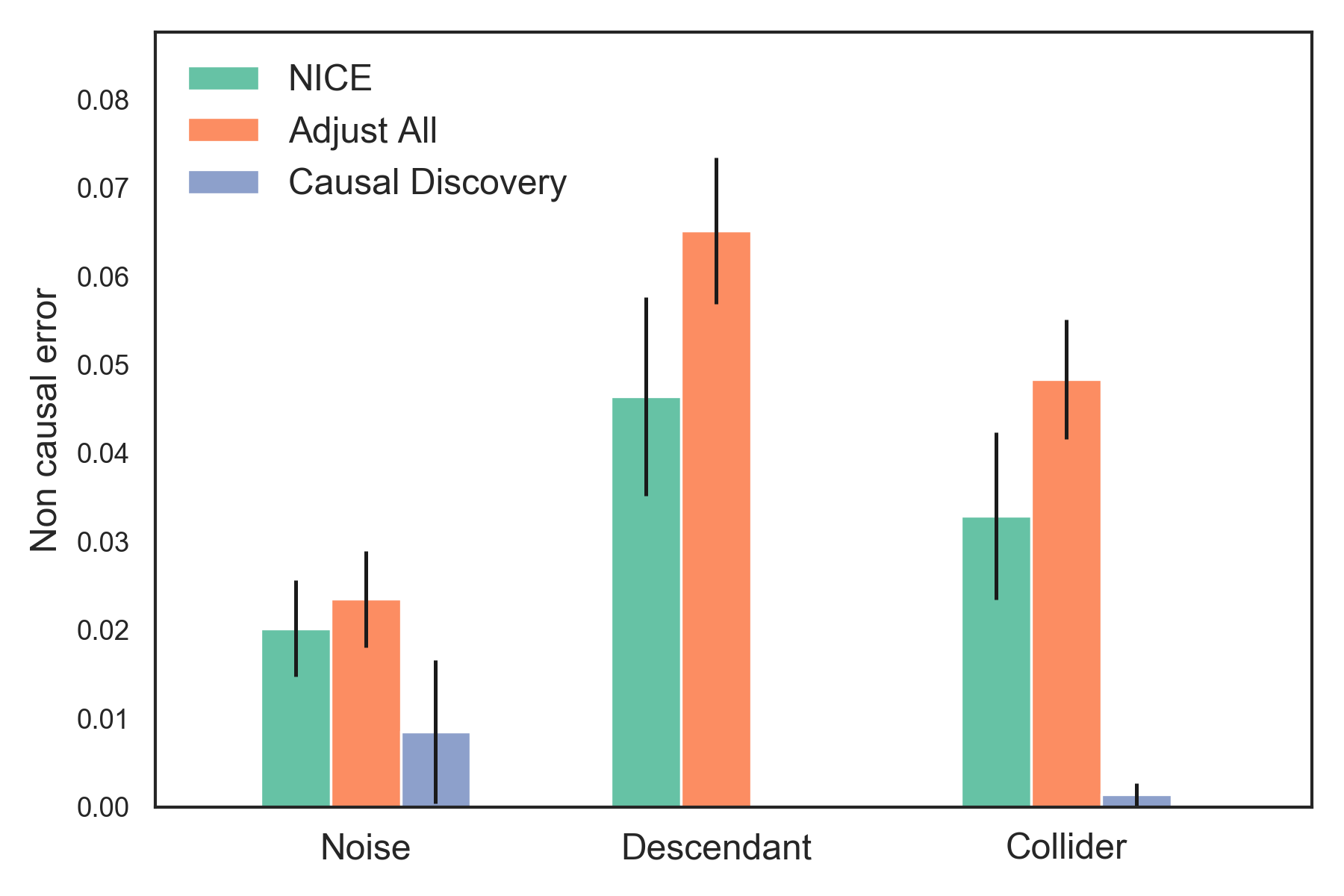}
    \caption{non causal error under the scrambled and heteroskedastic setting}\label{fig:exp-weight-1}

\end{figure}

\begin{figure}
      \centering
    \includegraphics[width=\linewidth]{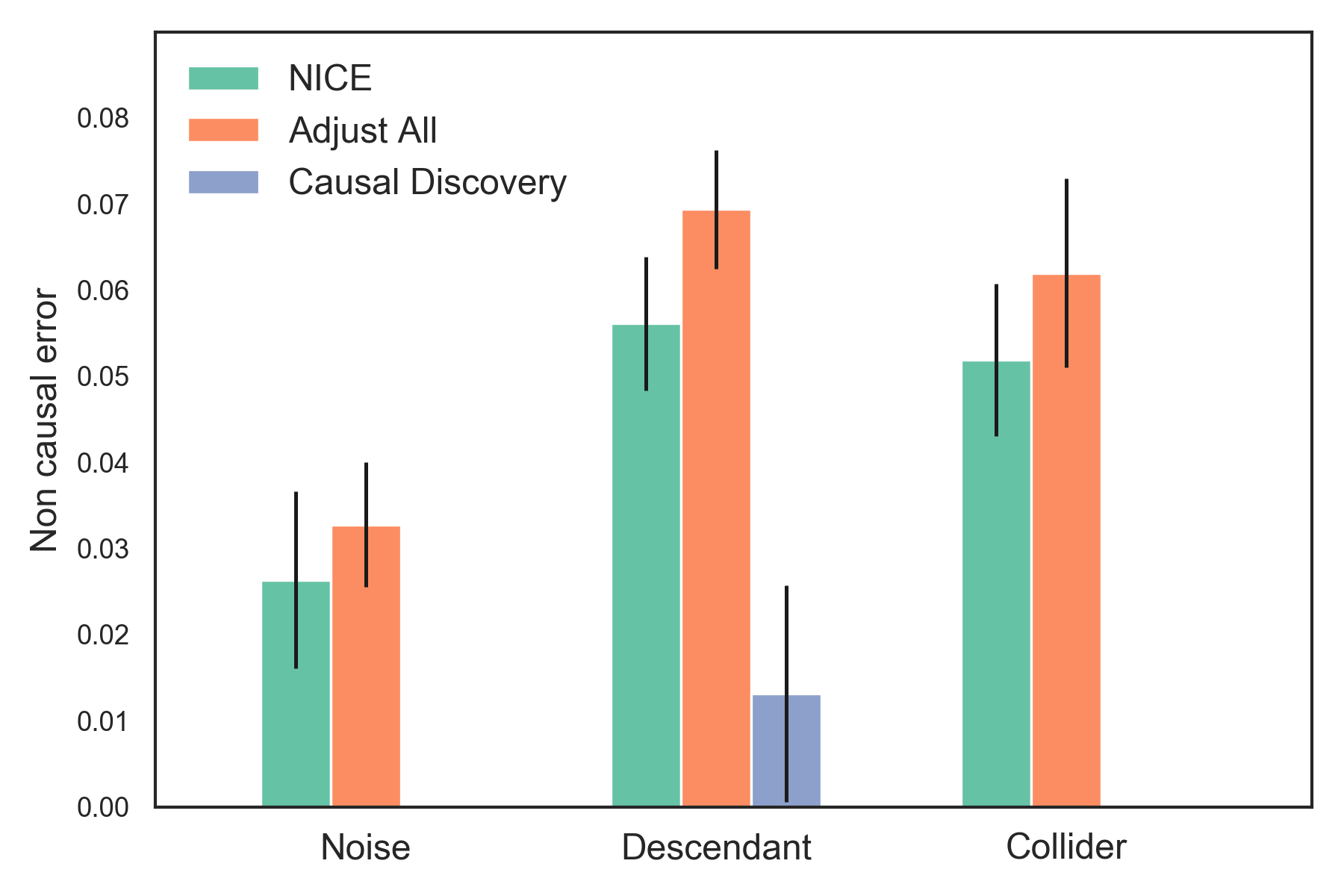}
    \caption{non causal error under the scrambled and homoscedastic setting}\label{fig:exp-weight-2}

\end{figure}

\begin{figure}
\centering
\includegraphics[width=\linewidth]{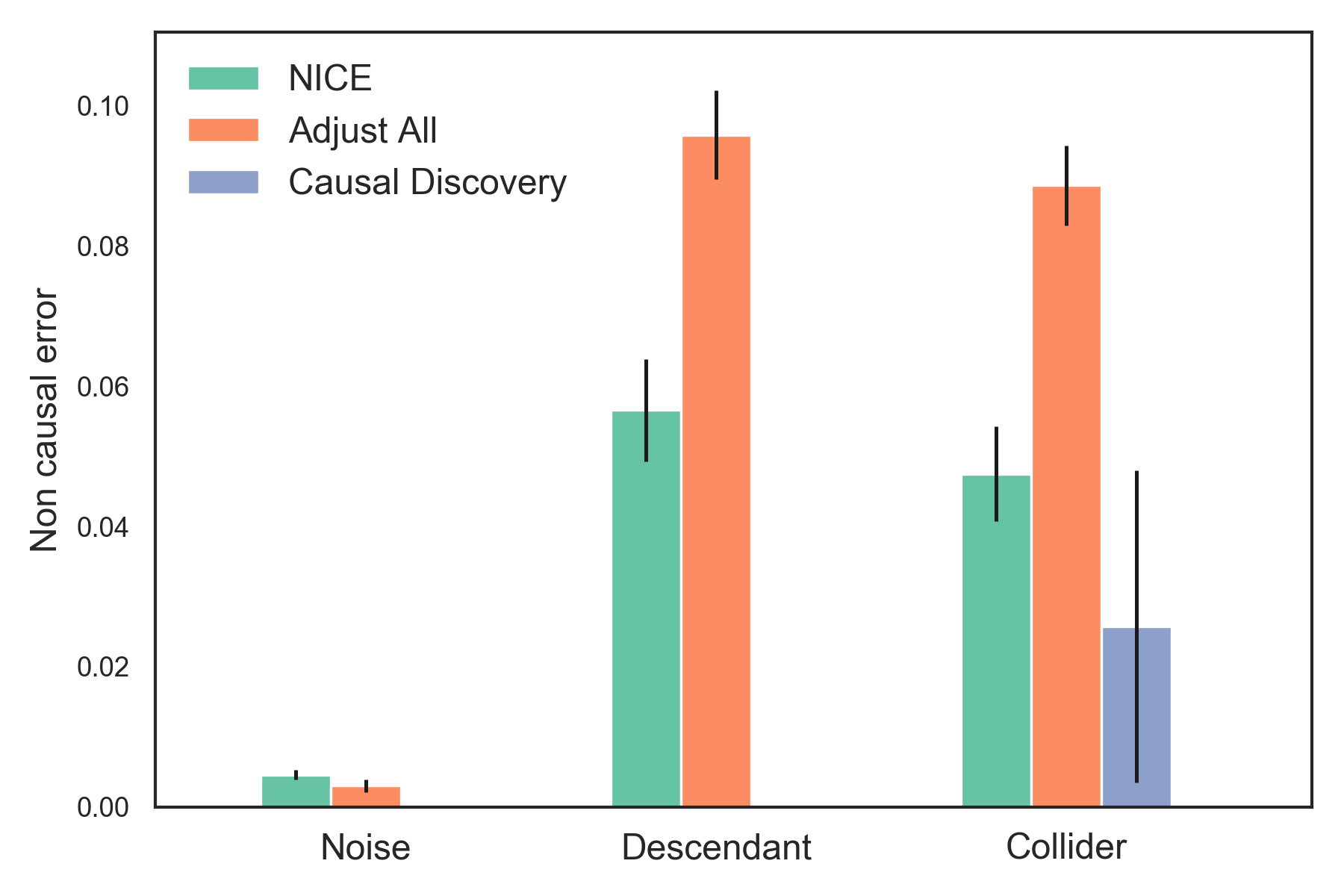}
\caption{non causal error under the unscrambled and heteroskedastic setting}\label{fig:exp-weight-3}
\end{figure}

\begin{figure}
  \centering
\includegraphics[width=\linewidth]{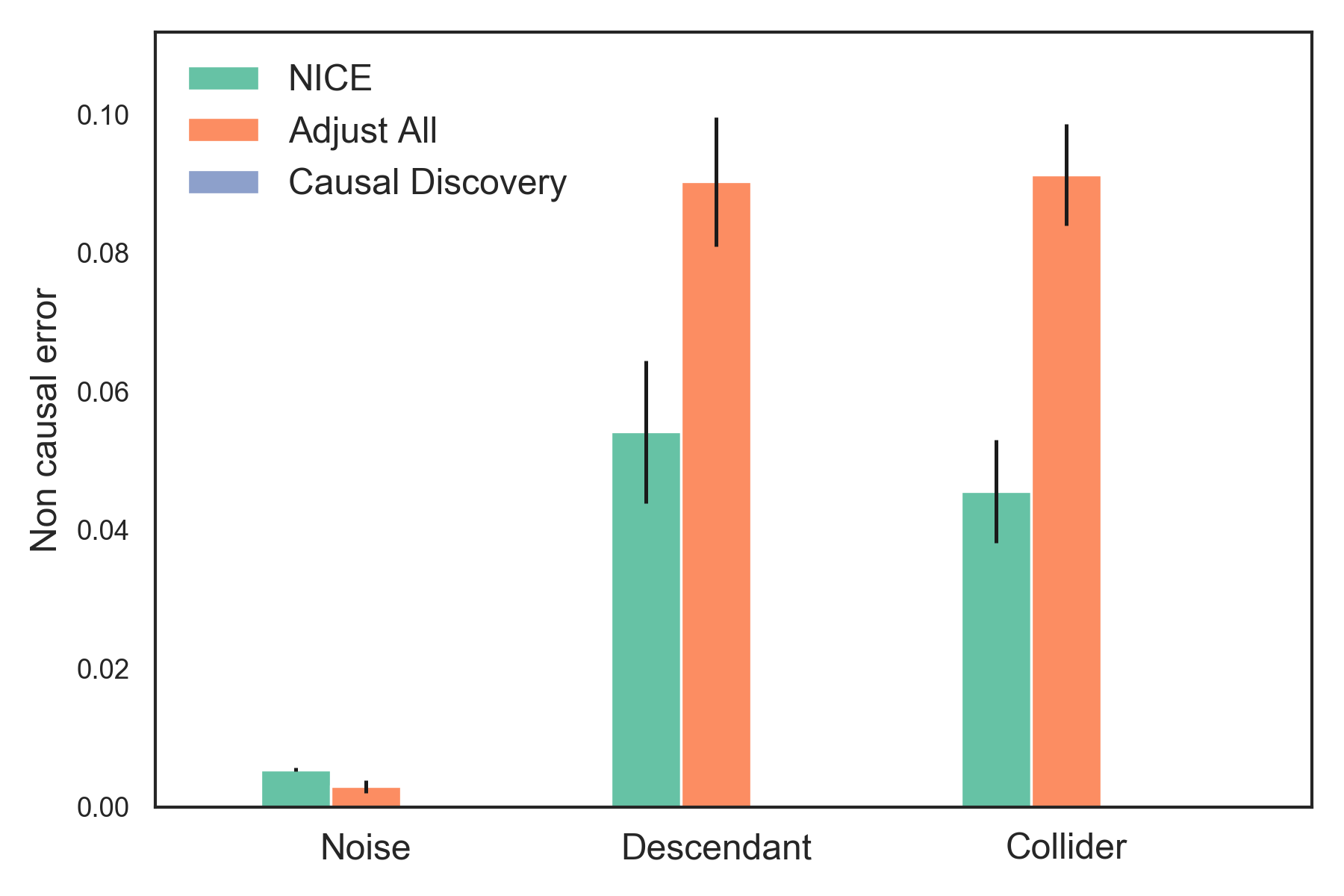}

\caption{non causal error under the scrambled and homoscedastic setting }\label{fig:exp-weight-4}
\end{figure}

To understand why NICE reduces the estimation bias, we measure the weight of the control predictor. The non-causal error is measured by the mean square error of the weight on $X_2$. The results are illustrated \Cref{fig:exp-weight-1}, \Cref{fig:exp-weight-2}, \Cref{fig:exp-weight-3}, and \Cref{fig:exp-weight-4}.



\subsubsection{Experiment 2}
We validate NICE for the non-linear case on a benchmark dataset, SpeedDating. SpeedDating was collected to study the gender difference in mate selection \citep{fisman2006gender}. The study recruited university students to participate in speed dating, and collected objective and subjective information such as `undergraduate institution' and `perceived attractiveness'. It has 8378 entries and 185 covariates. ACIC 2019's simulation samples subsets of the covariates to simulate binary treatment $T$ and binary outcome $Y$. Specifically, it provides four modified DGPs: Mod1: parametric models; Mod2: complex models; Mod3: parametric models with poor overlap; Mod4: complex models with treatment heterogeneity. Each modification includes three versions: low, med, high, indicating an increasing number of covariates included in the models for $T$ and $Y$.

We compare the estimation quality of the SATT and CATE over 10 bootstraps. We use two predictors: TARnet and Dragonnet. The main paper report the estimation quality of SATT using TARnet. We now report the estimation quality of CATE and SATT using Dragonnet, as well as CATE using TARnet.

The DGPs are recorded in the R file under SpeedDating folder. The original ACIC DGPs can be downloaded using the following link.
\begin{verbatim}
drive.google.com/file/d/
1Qqgmb3R9Vt9KTx6t8i_5IbFenylsPfrK/view
\end{verbatim}
We made several modifications to the original DGPs.
(1) ACIC competitions are usually designed to evaluate model performances using ATE or ATT. The simulation study provided ATE values but not ITE. In the SpeedDating DGPs, the treatment is binary, and the outcome is also binary. To calculate ITE, we take the difference between the propensity of the outcome of the treated predictor and the propensity of the outcome of the control predictor.
(2) Unfortunately, ACIC datasets do not come in multiple environments, nor do the covariates include bad controls.
To create multiple environments, we draw 6000 samples and select a covariate $x$ that's not the causal parent of $Y$. 
We sort the samples based on the the covariates value and divide them into three equal sized environments. 
For each DGP, we draw 10 bootstrap samples.

\Cref{tb:appendix-valid-att-dragon} and \Cref{tb:appendix-bad-att-dragon} show the average of the MAE of SATT estimates over three environments. The predictor uses the architecture of Dragonnet. We observe that NICE does not hurt the estimation quality in comparison to adjusting for all the covariates. 
When there is a strong collider in the adjustment set, NICE reduces collider bias across simulation setups.

\Cref{tb:appendix-valid-pehe-TARnet} and \Cref{tb:appendix-valid-pehe-dragon} show the PEHE of the CATE, with TARnet and dragonnet, respectively. We find that NICE improves the CATE estimates in Mod1, Mod2, and Mod3. This is surprising, as we expect NICE to perform equally well as adjusting for all covariates when the adjustment set is valid. To understand this phenomenon better, we examine the CATE estimates across simulation settings.

\begin{figure}
 \centering
 \includegraphics[width=\linewidth]{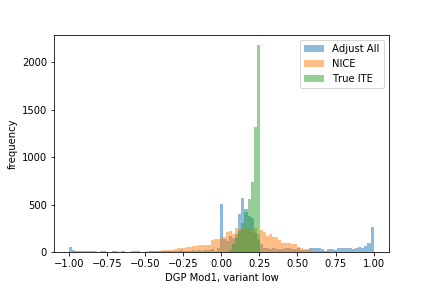}

\caption{ Mod1: parametric models }\label{fig:exp-mod1}
\end{figure}

\begin{figure}
  \centering
\includegraphics[width=\linewidth]{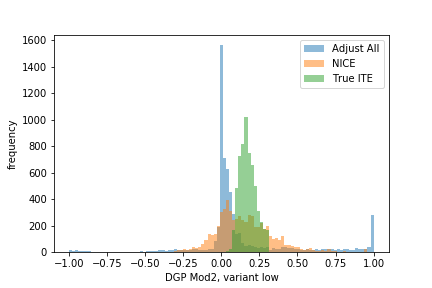}

\caption{Mod2: complex models}\label{fig:exp-mod2}
\end{figure}

\begin{figure}
     \centering
     \includegraphics[width=\linewidth]{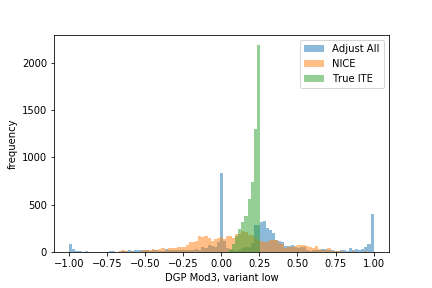}
    
    \caption{Mod1: parametric models with poor overlap}\label{fig:exp-mod3}

\end{figure}

\begin{figure}
      \centering
    \includegraphics[width=\linewidth]{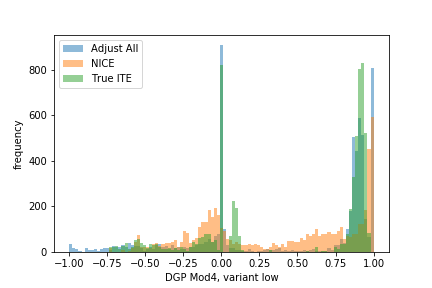}
    \caption{Mod4: complex models with treatment heterogeneity.}\label{fig:exp-mod4}

\end{figure}

In \Cref{fig:exp-mod1}, \Cref{fig:exp-mod2}, \Cref{fig:exp-mod3}, and \Cref{fig:exp-mod4} illustrate the estimation quality of CATE under the ``med" simulation setting. we compare the ground truth, the CATE estimates using NICE and Adjusting for all covariates across settings. Recall NICE uses an predictor trained with IRMv1 objective, adjusting for all uses a predictor that's trained with ERM objective. In Mod1, Mod2, and Mod3, there are little heterogeneity of the treatment effects. In Mod4,  the treatment effects are more spread out and heterogeneous.  
We observe that in Mod1, Mod2, and Mod3, the ERM predictors produce extreme CATE estimates. In contrast, using an IRMv1 objective, the CATE estimates are less extreme. In Mod4, where the CATE varies drastically, both IRMv1 and ERM predictors were able to capture the heterogeneity.

\begin{figure}
    \centering
    \includegraphics[width=\linewidth]{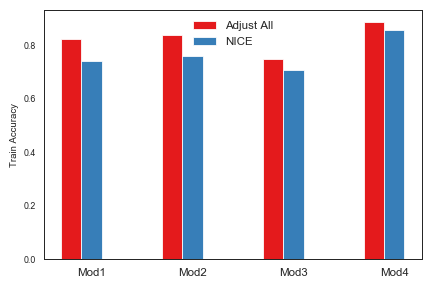}
    \caption{The training accuracy of the predictors}
    \label{fig:exp2-train-accuracy}
\end{figure}

\begin{figure}
    \centering
    \includegraphics[width=\linewidth]{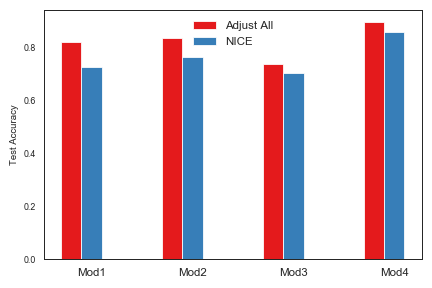}
    \caption{The testing accuracy of the predictors}
    \label{fig:exp2-test-accuracy}
\end{figure}

To examine whether the difference is due to over-fitting, we use two environments as training and one environment as testing. \Cref{fig:exp2-train-accuracy} and \Cref{fig:exp2-test-accuracy} show the corresponding training and testing accuracy across experiment setups. We observe the ERM predictors have similar training and testing accuracy. This suggests the model is not overfitting in the robust prediction sense.  We suspect that the IRMv1 penalty term becomes a regularization term that restrict the model to simpler solutions. Prior work \citep{janzing2019causal} has shown that regularizing terms in linear regression settings not only help against over-fitting finite data, but sometimes also produce better causal models in the infinite sample settings. There are no known results in the non-linear settings. 

Note that the ACIC competitions are \emph{not} designed for evaluating CATE performance. 
Estimating CATE when the outcomes are binary are difficult, especially given a flexible neural network model. 
The analysis above is about model specification for estimation, in settings \emph{without} bad controls.
In this paper, we consider the problem of causal adjustment. We focus on finding a causal representation that strips out bad controls.
Does an invariant predictor produce better causal estimate, even when there are no bad controls? We defer this question to future work. 

\begin{table}[ht]

\centering
  \resizebox{\linewidth}{!}{
\begin{tabular}{ll|cccc}

\multicolumn{3}{l}{valid adjustment}  & \multicolumn{3}{c}{$\epsilon_{pehe}$} \\
\hline
&TARnet & {\textsc \textbf{Mod1}} &  {\textsc\textbf{Mod2} }     & {\textsc \textbf{Mod3}} & {\textsc \textbf{Mod4}}\\
\hline
\textbf{low}  &Adjust All &$.16\pm .06$  & $.13\pm .05$ & $.22 \pm .18$ & $.05 \pm .01$\\
& NICE &$.06\pm .02$ & $.05\pm .02$  & $.06 \pm .02$ & $.05 \pm .01$\\
\hline
\textbf{med}& Adjust All &$.14 \pm .06$  & $.15\pm .02$ & $.12 \pm .02$ & $.06\pm .02$\\
& NICE &$.05\pm .01$ & $.05 \pm .01$  & $.07 \pm .03$ & $.04 \pm .01$\\
\hline
\textbf{high} & Adjust All&$.13\pm .07$  & $.12\pm .02$ & $.16 \pm .09$ & $.06\pm .01$\\
& NICE &$.05\pm .01$ & $.05 \pm .01$  & $.07\pm .02$ & $.04\pm .01$\\
\hline
\end{tabular}}

\caption{}\label{tb:appendix-valid-pehe-TARnet}
\end{table}

\begin{table}[ht]

\centering

  \resizebox{\linewidth}{!}{
\begin{tabular}{ll|cccc}

\multicolumn{3}{l}{bad controls in adjustment} & \multicolumn{3}{c}{$\epsilon_{pehe}$} \\
\hline
&TARnet & {\textsc \textbf{Mod1}} &  {\textsc\textbf{Mod2} }     & {\textsc \textbf{Mod3}} & {\textsc \textbf{Mod4}}\\
\hline
\textbf{low}  &Adjust All &$.35\pm .12$  & $.17\pm .02$ & $.30 \pm .04$ & $.24 \pm .04$\\
& NICE &$.08 \pm .04$ & $.05\pm .01$  & $.07 \pm .02$ & $.07 \pm .01$\\
\hline
\textbf{med}& Adjust All &$.26 \pm .02$  & $.27\pm .05$ & $.39 \pm .07$ & $.13\pm .01$\\
& NICE &$.04\pm .01$ & $.05\pm .02$  & $.06 \pm .01$ & $.04 \pm .01$\\
\hline
\textbf{high} & Adjust All&$.31 \pm .04$  & $.23\pm .06$ & $.31 \pm .07$ & $.12\pm .01$\\
& NICE &$.07\pm .02$ & $.07 \pm .03$  & $.14\pm .07$ & $.06\pm .03$\\
\hline
\end{tabular}}

\caption{}\label{tb:appendix-bad-pehe-TARnet}
\end{table}

\begin{table}[ht]
\centering

  \resizebox{\linewidth}{!}{
\begin{tabular}{ll|cccc}

\multicolumn{3}{l}{valid adjustment} & \multicolumn{3}{c}{$\epsilon_{att}$} \\
\hline
& Dragonnet & {\textsc \textbf{Mod1}} &  {\textsc\textbf{Mod2} }     & {\textsc \textbf{Mod3}} & {\textsc \textbf{Mod4}}\\
\hline
\textbf{low}  &Adjust All &$.09\pm .08$  & $.05\pm .03$ & $.04 \pm .04$ & $.03 \pm .01$\\
& NICE &$.06 \pm .03$ & $.03\pm .01$  & $.09 \pm .02$ & $.02 \pm .02$\\
\hline
\textbf{med}& Adjust All &$.06 \pm .09$  & $.06\pm .05$ & $.11 \pm .11$ & $.07\pm .05$\\
& NICE &$.07\pm .02$ & $.06\pm .03$  & $.08 \pm .03$ & $.04 \pm .03$\\
\hline
\textbf{high} & Adjust All&$.04 \pm .04$  & $.05\pm .08$ & $.03 \pm .02$ & $.02\pm .02$\\
& NICE &$.02\pm .01$ & $.07 \pm .02$  & $.06\pm .02$ & $.08\pm .05$\\
\hline
\end{tabular}}

\caption{}\label{tb:appendix-valid-att-dragon}
\end{table}

\begin{table}[ht]
\centering
  \resizebox{\linewidth}{!}{
\begin{tabular}{ll|cccc}
\multicolumn{3}{l}{bad controls in adjustment} & \multicolumn{3}{c}{$\epsilon_{att}$} \\
\hline
& Dragonnet & {\textsc \textbf{Mod1}} &  {\textsc\textbf{Mod2} }     & {\textsc \textbf{Mod3}} & {\textsc \textbf{Mod4}}\\
\hline
\textbf{low}  &Adjust All &$.31\pm .12$  & $.52\pm .16$ & $.39 \pm .06$ & $.75 \pm .11$\\
& NICE &$.17 \pm .10$ & $.03\pm .02$  & $.24 \pm .03$ & $.11 \pm .06$\\
\hline
\textbf{med}& Adjust All &$.54 \pm .09$  & $.47\pm .15$ & $.57 \pm .24$ & $.31\pm .13$\\
& NICE &$.13\pm .10$ & $.25\pm .05$  & $.15 \pm .08$ & $.05 \pm .03$\\
\hline
\textbf{high} & Adjust All &$.50\pm .06$  & $.58\pm .05$ & $.54 \pm .05$ & $.43 \pm .13$\\
& NICE &$.07 \pm .04$ & $.21\pm .06$  & $.17 \pm .05$ & $.08 \pm .03$\\
\hline
\end{tabular}}

\caption{}\label{tb:appendix-bad-att-dragon}
\end{table}

\begin{table}[ht]
\centering
  \resizebox{\linewidth}{!}{
\begin{tabular}{ll|cccc}

\multicolumn{3}{l}{valid adjustment}  & \multicolumn{3}{c}{$\epsilon_{pehe}$} \\
\hline
&Dragonnet & {\textsc \textbf{Mod1}} &  {\textsc\textbf{Mod2} }     & {\textsc \textbf{Mod3}} & {\textsc \textbf{Mod4}}\\
\hline
\textbf{low}  &Adjust All &$.20\pm .06$  & $.12\pm .02$ & $.19 \pm .02$ & $.06\pm .01$\\
& NICE &$.06\pm .01$ & $.04\pm .01$  & $.06 \pm .01$ & $.04 \pm .01$\\
\hline
\textbf{med}& Adjust All &$.15 \pm .06$  & $.16\pm .01$ & $.16 \pm .07$ & $.06\pm .01$\\
& NICE &$.04\pm .01$ & $.03\pm .01$  & $.07 \pm .02$ & $.03 \pm .01$\\
\hline
\textbf{high} & Adjust All &$.12\pm .02$  & $.14\pm .05$ & $.13 \pm .02$ & $.06 \pm .00$\\
& NICE &$.05\pm .01$ & $.06\pm .01$  & $.07 \pm .02$ & $.04 \pm .02$\\
\hline
\end{tabular}}
\caption{}\label{tb:appendix-valid-pehe-dragon}
\end{table}

\begin{table}[ht]
\centering
  \resizebox{\linewidth}{!}{
\begin{tabular}{ll|cccc}

\multicolumn{3}{l}{bad controls in adjustment}  & \multicolumn{3}{c}{$\epsilon_{pehe}$} \\
\hline
&Dragonnet & {\textsc \textbf{Mod1}} &  {\textsc\textbf{Mod2} }     & {\textsc \textbf{Mod3}} & {\textsc \textbf{Mod4}}\\
\hline
\textbf{low}  &Adjust All &$.47\pm .05$  & $.23\pm .05$ & $.37 \pm .05$ & $.46\pm .14$\\
& NICE &$.16\pm .08$ & $.05\pm .02$  & $.18 \pm .04$ & $.05 \pm .02$\\
\hline
\textbf{med}& Adjust All &$.32 \pm .05$  & $.37\pm .06$ & $.55 \pm .18$ & $.15\pm .02$\\
& NICE &$.07\pm .03$ & $.13\pm .04$  & $.11\pm .05$ & $.04 \pm .01$\\
\hline
\textbf{high} & Adjust All &$.38\pm .06$  & $.29\pm .05$ & $.39 \pm .05$ & $.14 \pm .02$\\
& NICE &$.09\pm .03$ & $.11\pm .04$  & $.13 \pm .05$ & $.04 \pm .01$\\
\hline
\end{tabular}}

\caption{}\label{tb:appendix-bad-pehe-dragon}
\end{table}

\subsubsection{Experiment 3}
In the third experiment, we examine the effect of environment variations on NICE's performance. 
We simulate non-linear data using the causal graph illustrated in \cref{fig:exp3-dgp}.
We draw three source environments $\{P^{e_1},P^{e_2}, P^{e_3}\}$, where $e_1 =0.2, e_2 = 1, e_3 =5$. In each source environments, we draw 900 samples. \Cref{fig:sample_porprotion} reports the average MAE of SATT over 5 simulations.  
\begin{align*}
A^e &\leftarrow \mathcal{N}(0, e^2)\\
X^e & \leftarrow A^e \cdot w_{ax^e}\\
X^e_t & \leftarrow X^e_{\{1...12\}}\\
X^e_y & \leftarrow X^e_{\{13...30\}}\\
p^e_t &\leftarrow \text{sigmoid} (f(X^e_t))\\
T^e &\leftarrow Bern( p^e_t)\\
p^e_y &\leftarrow \text{sigmoid}(g(X^e_t,X^e_y, T^e))\\
Y^e &\leftarrow B(n, p^e_y)\\
Z^e &\leftarrow  Y^e+ T^e +\mathcal{N}(0,1)
\end{align*}
$f(X^e_t) = X^e_t \cdot w_{xt^e} + h(X_t^e)\cdot w_{xt^{e'}} $, where $h(X_t^e)$ is implemented as \begin{verbatim}
  [X_t[:, :1] * X_t[:, 1:2],
  X_t[:, 1:2] * X_t[:, 2:4],
  X_t[:, 2:3] * X_t[:, 3:] 
  / np.square(X_t).mean()]
\end{verbatim}
$g(X^e_t,X^e_y, T^e) = 1.25*T^e + X_t\cdot w_{xy^e} +2* p^e_t + m(x^e_y) \cdot w_{xy^{e'}} $

Here $m(x^e_y) $ is implemented as
\begin{verbatim}
    [X_y[:, :1] * X_y[:, 4:5], 
    X_y[:, 1:2] * X_y[:, 3:4],
    X_y[:, 1:2] * X_y[:, 2:] 
    / np.square(X_y).mean()]
\end{verbatim}
The complete data generating code is under \begin{verbatim}diverse_environments/gen_dat.py\end{verbatim}
New environments $P^{e'_1}, P^{e'_2}, P^{e'_3}$  are mixtures of the three source environments $P^{e_1}, P^{e_2}, P^{e_3}$ .
Respectively, $P^{e'_1}, P^{e'_2}, P^{e'_3}$ draw ($p_1$, $p_2$, $p_3$) proportions from from $P^{e_1}$, ($p_2$, $p_3$, $p_1$) proportions from from $P^{e_2}$, and ($p_2$, $p_3$, $p_1$) proportions from from $P^{e_3}$.The proportions $(p_1$, $p_2$, $p_3)$ sum to one.

The mixing proportions we considered are: $(0,0,1)$,  $(0,0.1,0.9)$, $(0,0.2,0.8)$, $(0,0.3,0.7)$, $(0,0.4,0.6)$, $(0,0.5,0.5)$,$(0.1,0.1,0.8)$,$(0.1,0.2,0.7)$,$(0.1, 0.3, 0.6)$, $(0.1, 0.4, 0.5)$,$(0.2, 0.2, 0.6)$, $(0.2, 0.3, 0.5)$,$(0.2, 0.4, 0.4)$, $(0.3, 0.3, 0.4)$.



\end{document}
